\documentclass[runningheads]{llncs}
\usepackage[table,xcdraw]{xcolor}
 
 \usepackage{eccv}

\usepackage{eccvabbrv}

\usepackage{graphicx}
\usepackage{booktabs}
\usepackage{amsmath}
\usepackage{amssymb}
\usepackage{multirow}
\usepackage{wrapfig}
\usepackage[accsupp]{axessibility}  

 \usepackage{hyperref}

\usepackage{orcidlink}

\definecolor{c2}{HTML}{FBD9BD}
\definecolor{c3}{HTML}{fe793d}
\definecolor{c4}{HTML}{eedeb0}
\definecolor{rouse}{rgb}{0.981,0.961,0.941}

\makeatletter
\newcommand*\bigcdot{\mathpalette\bigcdot@{.5}}
\newcommand*\bigcdot@[2]{\mathbin{\vcenter{\hbox{\scalebox{#2}{$\m@th#1\bullet$}}}}}
\makeatother

\begin{document}

\title{Reti-Diff: Illumination Degradation Image Restoration with Retinex-based \\ Latent Diffusion Model} 

\titlerunning{Reti-Diff}

\author{Chunming He$^{1,*}$\,,
        Chengyu Fang$^{1,}$\thanks{Equal Contribution, $\dagger$ Corresponding Author}\,~,
        Yulun Zhang$^{2,\dagger}$ \,, 
        Tian Ye$^{3}$ \,, \\
	Kai Li$^{4}$\,,
	Longxiang Tang$^1$\,,
	Zhenhua Guo$^5$\,,
	Xiu Li$^{1,\dagger}$\,,
        and Sina Farsiu$^6$\\
 }

\authorrunning{He \& Fang et al.}

\institute{Shenzhen International Graduate School, Tsinghua University \and
Shanghai Jiao Tong University \and
The Hong Kong University of Science and Technology (Guangzhou) \and NEC Laboratories America \and Tianyi Traffic Technology \and Duke University \\
}

\maketitle

\begin{abstract}
  Illumination degradation image restoration (IDIR) techniques aim to improve the visibility of degraded images and mitigate the adverse effects of deteriorated illumination. Among these algorithms, diffusion model (DM)-based methods have shown promising performance but are often burdened by heavy computational demands and pixel misalignment issues when predicting the image-level distribution. To tackle these problems, we propose to leverage DM within a compact latent space to generate concise guidance priors and introduce a novel solution called Reti-Diff for the IDIR task. Reti-Diff comprises two key components: the Retinex-based latent DM (RLDM) and the Retinex-guided transformer (RGformer). To ensure detailed reconstruction and illumination correction, RLDM is empowered to acquire Retinex knowledge and extract reflectance and illumination priors. These priors are subsequently utilized by RGformer to guide the decomposition of image features into their respective reflectance and illumination components. Following this, RGformer further enhances and consolidates the decomposed features, resulting in the production of refined images with consistent content and robustness to handle complex degradation scenarios. Extensive experiments show that Reti-Diff outperforms existing methods on three IDIR tasks, as well as downstream applications. The code will be released.
\end{abstract}

\setlength{\abovedisplayskip}{2pt}
\setlength{\belowdisplayskip}{2pt}

\section{Introduction}\label{sec:intro}
Illumination degradation image restoration (IDIR) seeks to enhance the visibility and contrast of degraded images while mitigating the adverse effects of deteriorated illumination, \eg, indefinite noise and variable color deviation. IDIR has been investigated in various domains, including low-light image enhancement~\cite{Cai2023}, underwater image enhancement~\cite{guo2023underwater}, and backlit image enhancement~\cite{liang2023iterative}. 
By addressing illumination degradation, the enhanced images are expected to exhibit improved visual quality, making them more suitable for decision-making or subsequent tasks like nighttime object detection and segmentation.

\begin{figure}[t]
	\centering
	\begin{subfigure}{0.236\textwidth}
		\centering
		\includegraphics[width=\textwidth]{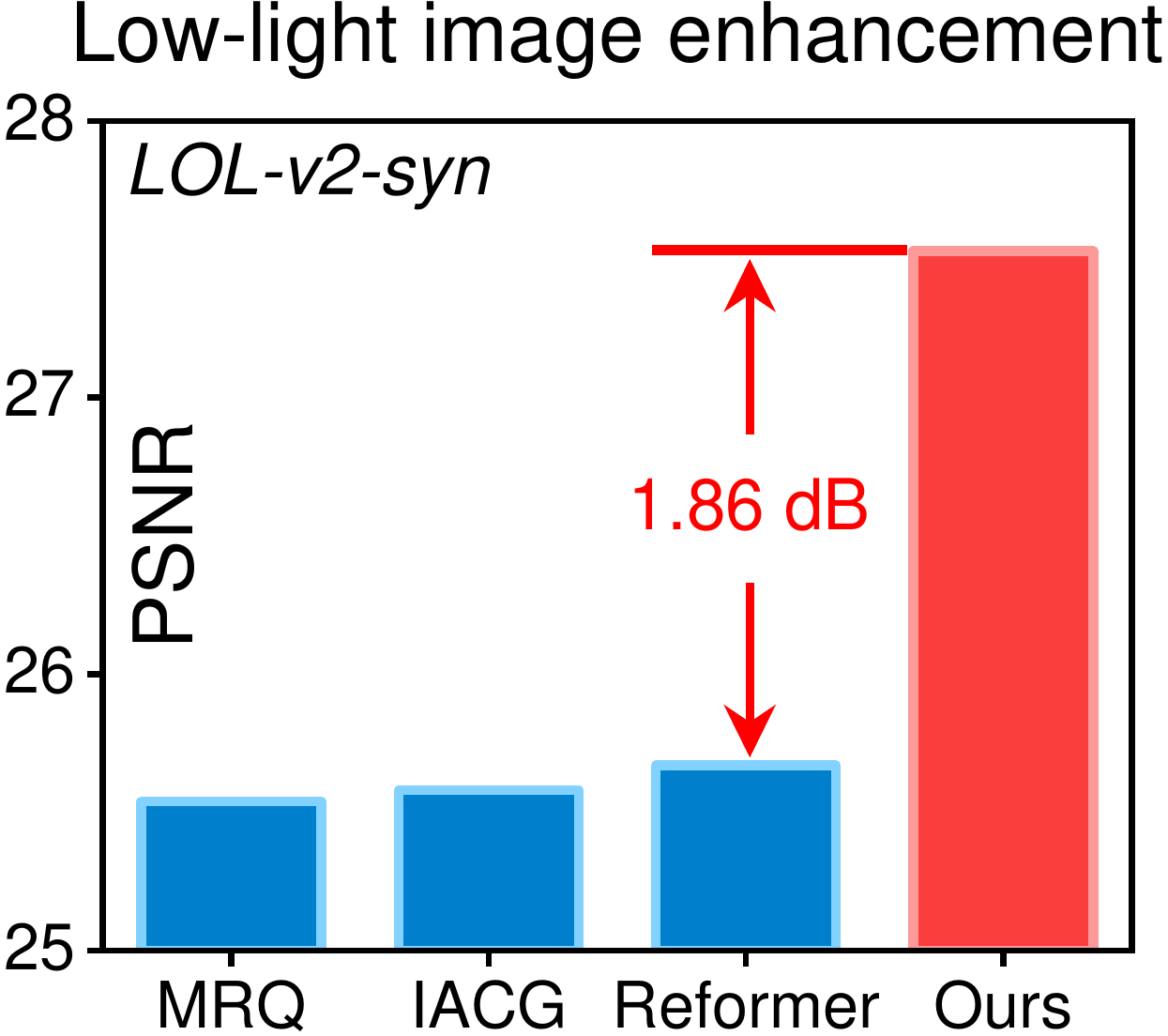}
	\end{subfigure}
	\hfill
	\begin{subfigure}{0.245\textwidth}  
		\centering 
		\includegraphics[width=\textwidth]{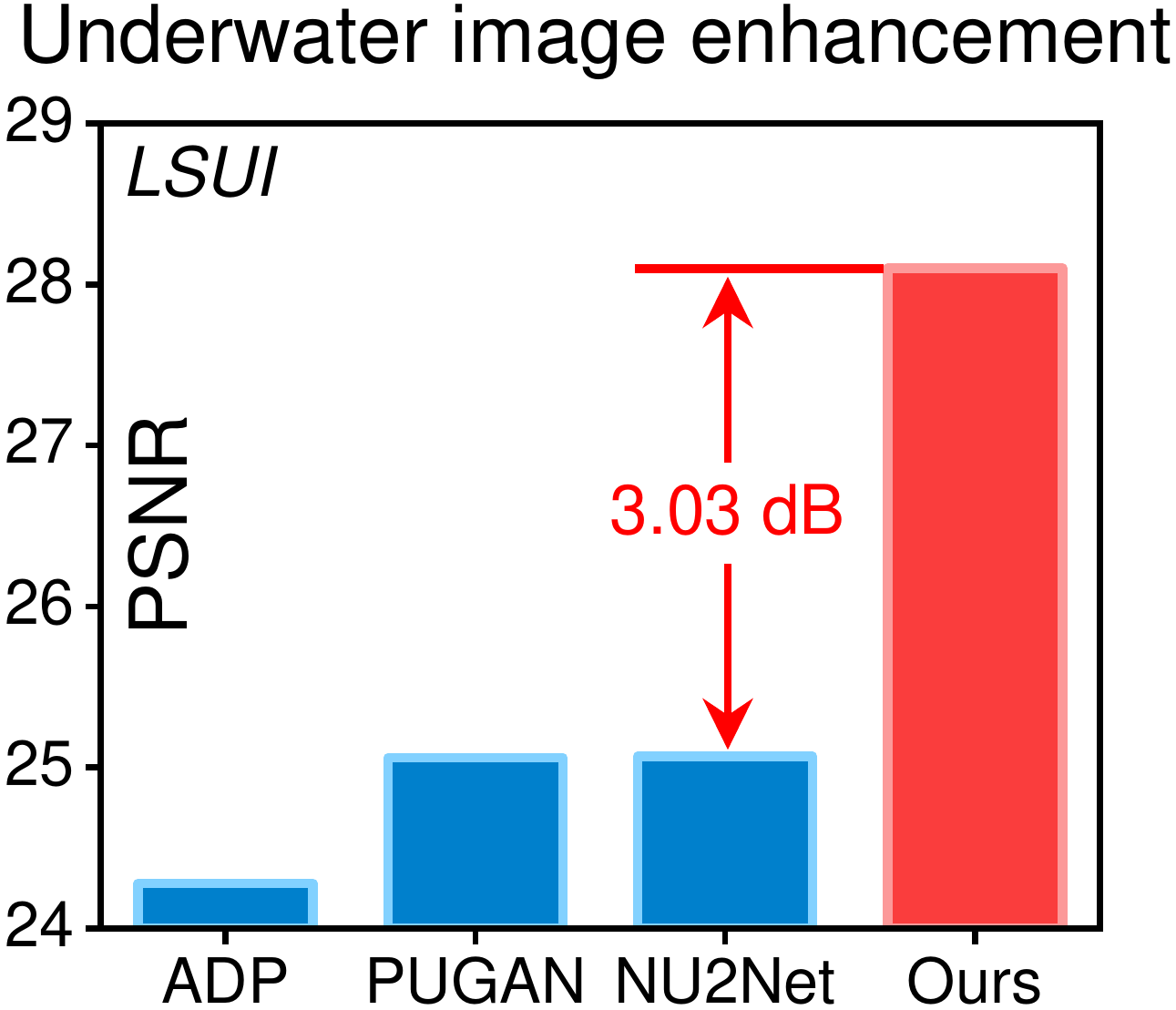}
	\end{subfigure}
    \begin{subfigure}{0.231\textwidth}
		\centering
		\includegraphics[width=\textwidth]{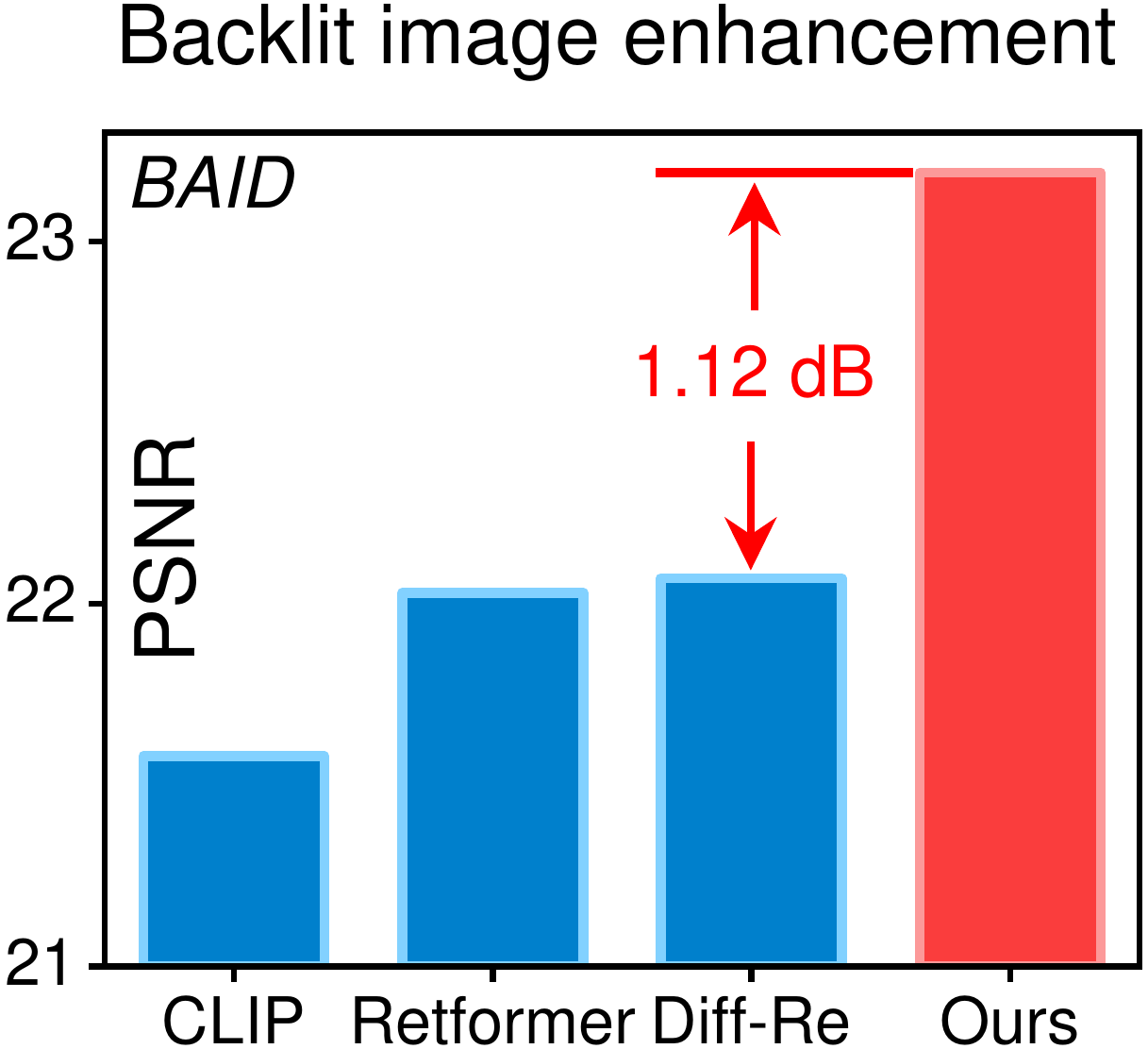}
	\end{subfigure}
	\hfill
	\begin{subfigure}{0.23\textwidth}  
		\centering 
		\includegraphics[width=\textwidth]{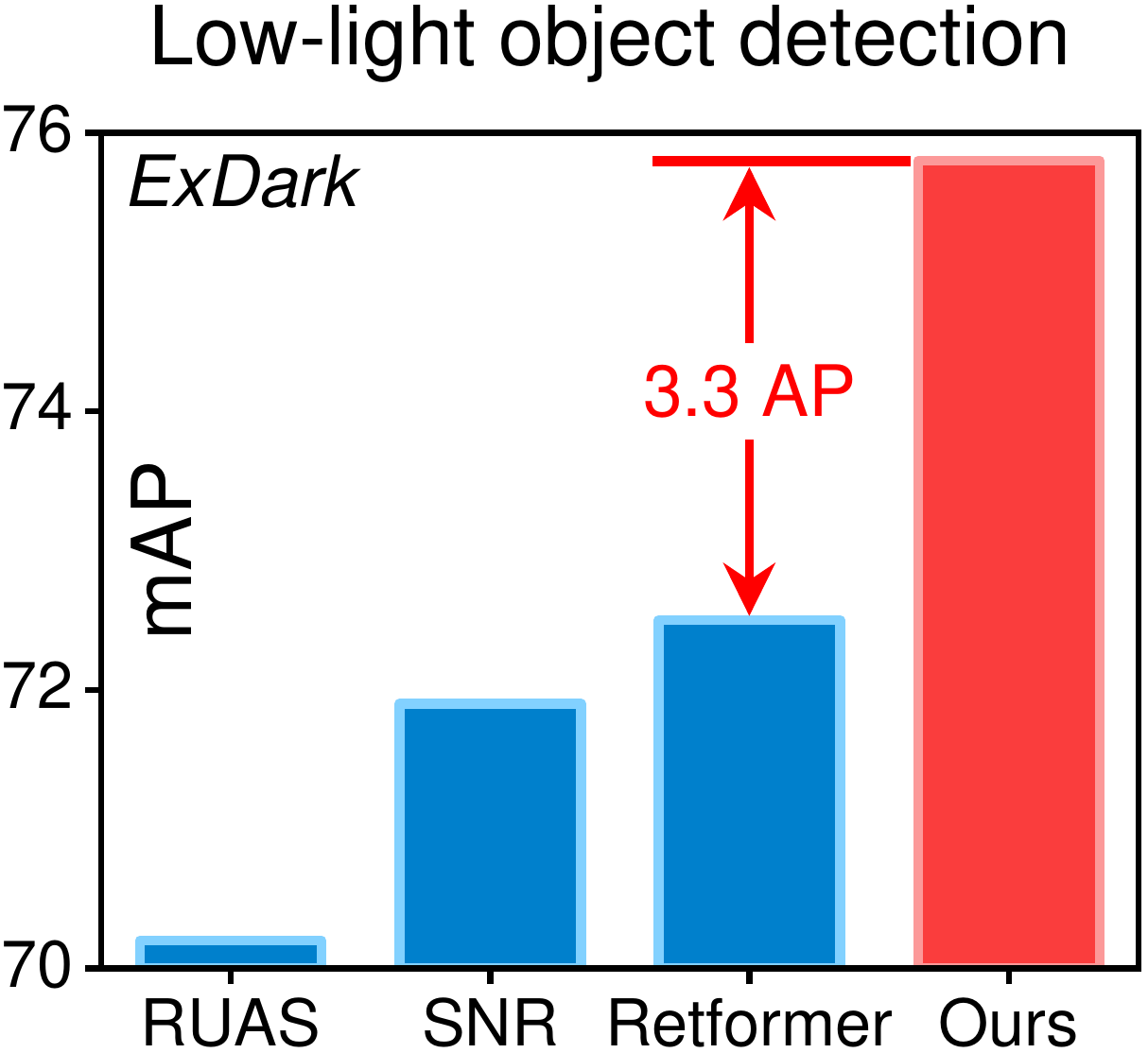}
	\end{subfigure}
\vspace{-3mm}
	\caption{Our Reti-Diff outperforms cutting-edge techniques on three IDIR tasks and the low-light object detection task, where CLIP, Diff-Re, and SNR are short for CLIP-LIT~\cite{liang2023iterative}, Diff-Retinex~\cite{yi2023diff}, and SNR-Net~\cite{xu2022snr}.
 }
	\label{fig:Superiority}
	\vspace{-7mm}
\end{figure}

Traditional IDIR approaches~\cite{fu2016fusion,ueng1995gamma} primarily rely on manually crafted enhancement techniques with limited generalization capabilities. Leveraging the robust feature extraction capabilities of convolutional neural networks and transformers, a series of deep learning-based methods~\cite{Cai2023,jiang2021enlightengan} have been proposed and have achieved remarkable success in the IDIR domain. However, 
as depicted in~\cref{fig:Superiority}, 
they still face challenges in complex illumination degradation scenarios due to their constrained generative capacity.

To overcome these challenges, deep generative models, like generative adversarial networks~\cite{he2023hqg} and variational autoencoder~\cite{he2024strategic},
have gained popularity in the IDIR task for their generative abilities. Recently, the diffusion model (DM)~\cite{yi2023diff} has been introduced to the IDIR field for high-quality image restoration.
However, existing DM-based methods, \eg, Diff-Retinex~\cite{yi2023diff} and GSAD~\cite{jinhui2023global}, apply DM directly to image-level generation, leading to two main challenges: \textbf{(1)} These methods incur high computational costs, as predicting the image-level distribution requires a large number of inference steps. \textbf{(2)} The enhanced results may exhibit pixel misalignment with the original clean image in terms of restored details and local consistency. 


To tackle the above problems, we propose introducing the latent diffusion model (LDM) to solve the IDIR problem. By applying DM in the low-dimensional compact latent space, we effectively alleviate the computational burden. Additionally, we incorporate LDM into transformers to prevent pixel misalignment in the generated image, which is often observed in existing deep generative models. 
Unlike existing LDM-based methods that solely use the priors extracted from the RGB domain, our method, tailored to the specific characteristics of IDIR tasks, empowers LDMs to extract Retinex information from both the reflectance and illumination domains.
This adaptation allows our method to generate high-fidelity Retinex priors directly from low-quality input images. By doing so, this approach enables us to simultaneously enhance image details using the reflectance prior and correct color distortions with the illumination prior, resulting in visually appealing results with favorable downstream tasks.



With this inspiration, we present Reti-Diff, the first LDM-based solution to tackle the IDIR problem. Reti-Diff, depicted in~\cref{fig:framework}, consists of two primary components: the Retinex-based LDM (RLDM) and the Retinex-guided transformer (RGformer). Initially, RLDM is employed to generate Retinex priors, which are then integrated into RGformer to produce visually appealing results. To ensure the generation of high-quality priors, we propose a two-phase training approach, wherein Reti-Diff undergoes initial pretraining followed by subsequent RLDM optimization.
\textbf{In phase \uppercase\expandafter{\romannumeral1}}, we introduce a Retinex prior extraction (RPE) module to compress the ground-truth image into the highly compact Retinex priors, namely the reflectance prior and the illumination prior. These priors are then sent to RGformer to guide feature decomposition and the generation of reflectance and illumination features. Afterward, RGformer employs the Retinex-guided multi-head cross attention (RG-MCA) and dynamic feature aggregation (DFA) module to refine and aggregate the decomposed features, ultimately producing enhanced images with coherent content and ensuring robustness and generalization in extreme degradation scenarios. \textbf{In phase \uppercase\expandafter{\romannumeral2}}, we train RLDM in reflectance and illumination domains to estimate Retinex priors from the low-quality image, with the constraint of consistency with those extracted by RPE from the ground-truth image. Therefore, the extracted Retinex priors can guide the RGformer in detail enhancement and illumination correction, resulting in visually appealing results with favorable downstream performance.

Our contributions are summarized as follows:








\begin{itemize}
    \vspace{-6pt}
	\item[$\bullet$] We propose a novel DM-based framework, Reti-Diff, for the IDIR task. To the best of our knowledge, this is the first application of the latent diffusion model to tackle the IDIR problem.

	\item[$\bullet$] 
We propose to let RLDM learn Retinex knowledge and generate high-quality reflectance and illumination priors from the low-quality input, which serve as critical guidance in detail enhancement and illumination correction.
 
	\item[$\bullet$] 
We propose RGformer to integrate extracted Retinex priors to decompose features into reflectance and illumination components and then utilize RG-MCA and DFA to refine and aggregate the decomposed features, ensuring robustness and generalization in complex illumination degradation scenarios.

 
	\item[$\bullet$] Extensive experiments on three IDIR tasks verify our superiority to existing methods in terms of image quality and favorability in downstream applications, including low-light object detection and segmentation. 
\end{itemize}

\section{Related Works}
\label{sec:RelWork}
 
\noindent\textbf{Illumination Degradation Image Restoration.}
Early IDIR methods mainly include three approaches: histogram equalization (HE)~\cite{abdullah2007dynamic}, gamma correction (GC)~\cite{huang2012efficient}, and Retinex theory~\cite{land1977retinex}. HE-based and GC-based methods focus on directly amplifying the low
contrast regions
but overlook illumination factors. Retinex-based variants~\cite{fu2016weighted,li2018structure} propose the development of priors to constrain the solution space for reflectance and illumination maps. However, these methods still rely on hand-crafted priors, limiting their ability to generalize effectively.
With the rapid development of deep learning, approaches based on CNNs and transformers~\cite{Cai2023,jiang2021enlightengan,he2023hqg} have achieved remarkable success in IDIR. For instance, LLNet~\cite{lore2017llnet} proposed a sparse denoising structure to enhance illumination and suppress noise. DIE~\cite{wang2019underexposed} integrated Retinex cues into a learning-based structure, presenting a one-stage Retinex-based solution for color correction. To enhance generative capacity, Diff-Retinex~\cite{yi2023diff} and GSAD~\cite{jinhui2023global} introduced DM to the IDIR field by directly applying it to image-level generation. However, they entail significant computational costs and may lead to pixel misalignment with the original input, particularly concerning restored image details and local consistency.

\noindent\textbf{Diffusion Models.}  
Diffusion models (DMs) have demonstrated considerable success in various domains, including density estimation~\cite{kingma2021variational} and data generation~\cite{he2024strategic}. Such a probabilistic generative model adopts a parameterized Markov chain to optimize the lower variational bound on the likelihood function, enabling them to generate target distributions with greater accuracy than other generative models, \ie, GAN and VAE. Recently, DMs have been introduced to solve the IDIR problem~\cite{yi2023diff,jinhui2023global}. However, when directly applied to image-level generation, these approaches introduce computational burdens and pixel misalignment. To overcome this, we propose employing LDM to estimate priors within a low-dimensional latent space. We then integrate these priors into the transformer-based framework, thus addressing the above problems. Besides, unlike existing LDM-based methods~\cite{xia2023diffir,chen2023hierarchical} that solely rely on priors extracted from the RGB domain, our method, tailored to the specific characteristics of IDIR tasks, empowers LDMs to extract Retinex information from both the reflectance and illumination domains.
This adaptation allows our method to generate high-fidelity Retinex priors directly from low-quality input images. By doing so, this novel approach enables us to simultaneously enhance image details using the reflectance prior and correct color distortions with the illumination prior, resulting in visually appealing results with favorable downstream tasks.

\vspace{-1mm}
\section{Methodology}\vspace{-1mm}
In this paper, we propose Reti-Diff, the pioneering method based on Latent Diffusion Models (LDM) for IDIR tasks. Reti-Diff is specifically tailored to address the challenges inherent in IDIR tasks by leveraging Retinex priors extracted from both the illumination and reflectance domains to guide the restoration process. This innovative approach utilizes the extracted Retinex prior representation as dynamic modulation parameters, facilitating simultaneous enhancement of restoration details through the reflectance prior and correction of color distortion via the illumination prior. This ensures the generation of visually compelling results while positively impacting downstream tasks.
\begin{figure*}[t]
	\centering
	\setlength{\abovecaptionskip}{-0.2cm}
	\begin{center}
		\includegraphics[width=\linewidth]{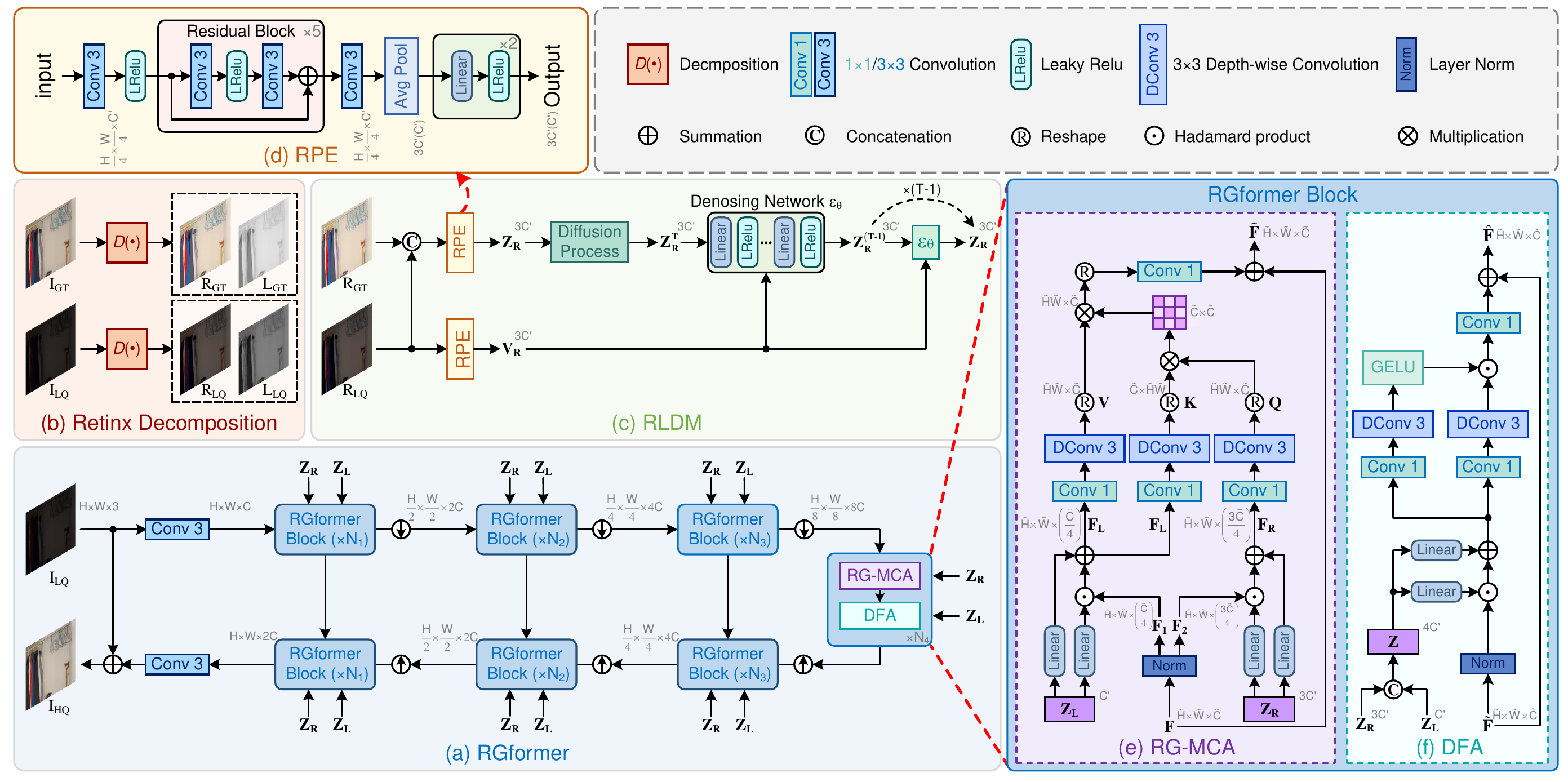}
	\end{center}
	\caption{Framework of Reti-Diff. We first pretrain Reti-Diff to ensure the robust learning of RLDM and then optimize RLDM to generate high-quality Retinex priors, which guide RGformer in detail enhancement and illumination correction.
In (a),
we omit the auxiliary decoder $D_a(\bigcdot)$ for simplicity.
    In (c), we only give the example by using RLDM to extract the reflectance prior and the illumination prior can be extracted similarly. }
	\label{fig:framework}
	\vspace{-0.6cm}
\end{figure*}

As shown in~\cref{fig:framework}, our Reti-Diff comprises two parts: the Retinex-guided transformer (RGformer) and the Retinex-based latent diffusion model (RLDM). 
To ensure the generation of high-quality priors, Reti-Diff undergoes a two-phase training strategy, involving the initial pretraining of Reti-Diff and the subsequent optimization of RLDM. In this section, we provide an in-depth explanation of the two-phase training approach and elucidate the entire restoration process.



\vspace{-2mm}
\subsection{Pretrain Reti-Diff}\vspace{-2mm}
We first pretrain Reti-Diff to encode the clear image, termed ground truth, into compact priors with Retinex prior extraction (RPE) module and use the extracted Retinex priors to guide RGformer for restoration. 

\noindent\textbf{Retinex prior extraction module.}
Given the low-quality (LQ) image $\mathbf{I}_{LQ}\in \mathbb{R}^{H\times W \times3}$ and its corresponding ground truth $\mathbf{I}_{GT}\in \mathbb{R}^{H\times W \times3}$, we initially decompose them into the reflectance image $\mathbf{R}\in \mathbb{R}^{H\times W \times3}$ and the illumination map $\mathbf{L}\in \mathbb{R}^{H\times W}$ according to Retinex theory:
\begin{equation}
    \mathbf{I}_{LQ}=\mathbf{R}_{LQ} \odot \mathbf{L}_{LQ}, \mathbf{I}_{GT}=\mathbf{R}_{GT} \odot \mathbf{L}_{GT},
\end{equation}
where $\odot$ denotes the Hadamard product. Following Retformer~\cite{Cai2023}, We use a pretrained decomposing network $D(\bigcdot)$ to decompose $\mathbf{I}_{LQ}$ and $\mathbf{I}_{GT}$. Then we concatenate
the corresponding components of ground truth and LQ image and use the RPE module $\text{RPE}(\bigcdot)$ to encode them into Retinex priors $\mathbf{Z}_{\mathbf{R}}\in \mathbb{R}^{3C'}$, $\mathbf{Z}_{\mathbf{L}}\in \mathbb{R}^{C'}$:
\begin{equation}
\setlength{\abovedisplayskip}{-5pt}
\setlength{\belowdisplayskip}{-5pt}
\begin{aligned}
&\mathbf{Z}_{\mathbf{R}}=\text{RPE}(\text{down}(\text{conca}(\mathbf{R}_{GT},\mathbf{R}_{LQ}))), \\
&\mathbf{Z}_{\mathbf{L}}=\text{RPE}(\text{down}(\text{conca}(\mathbf{L}_{GT},\mathbf{L}_{LQ}))),
\end{aligned}
\end{equation}
where $\text{conca}(\bigcdot)$ is concatenation and $\text{down}(\bigcdot)$ is downsampling that is operated by PixelUnshuffle. We then send Retinex priors, $\mathbf{Z}_{\mathbf{R}}$ and $\mathbf{Z}_{\mathbf{L}}$, to RGformer to serve as dynamic modulation parameters for detail restoration and color correction.


\noindent\textbf{Retinex-guided transformer.} 
RGformer mainly consists of two parts in each block, \textit{i.e.}, Retinex-guided multi-head cross attention (RG-MCA) and dynamic feature aggregation (DFA) module. In RG-MCA, we first split the input feature $\mathbf{F}\in\mathbb{R}^{\tilde{H}\times \tilde{W} \times \tilde{C}}$ into two parts $\mathbf{F}_1\in\mathbb{R}^{\tilde{H}\times \tilde{W} \times (3\tilde{C}/4)}$ and $\mathbf{F}_2\in\mathbb{R}^{\tilde{H}\times \tilde{W} \times (\tilde{C}/4)}$ along the channel dimension. Afterwards, we integrated $\mathbf{Z}_{\mathbf{R}}$ and $\mathbf{Z}_{\mathbf{L}}$ as the corresponding dynamic modulation parameters to generate reflectance-guided feature $\mathbf{F}_{\mathbf{R}}\in\mathbb{R}^{\tilde{H}\times \tilde{W} \times (3\tilde{C}/4)}$ and illumination-guided feature $\mathbf{F}_{\mathbf{L}}\in\mathbb{R}^{\tilde{H}\times \tilde{W} \times (\tilde{C}/4)}$:
\begin{equation}
\begin{aligned}
    &\mathbf{F}_{\mathbf{R}}=\text{Li}_1 (\mathbf{Z}_{\mathbf{R}})\odot \text{Norm}(\mathbf{F}_1)+\text{Li}_2 (\mathbf{Z}_{\mathbf{R}}), \\
    &\mathbf{F}_{\mathbf{L}}=\text{Li}_1 (\mathbf{Z}_{\mathbf{L}})\odot \text{Norm}(\mathbf{F}_2)+\text{Li}_2 (\mathbf{Z}_{\mathbf{L}}),
\end{aligned}
\end{equation}
where $\text{Norm}(\bigcdot)$ is layer normalization. $\text{Li}(\bigcdot)$ means linear layer. 
Afterward, we aggregate global spatial information by projecting $\mathbf{F}_{\mathbf{R}}$ into query $\mathbf{Q}=\mathbf{W}_Q \mathbf{F}_{\mathbf{R}}$ and key $\mathbf{K}=\mathbf{W}_K \mathbf{F}_{\mathbf{L}}$ and transforming $\mathbf{F}_{\mathbf{L}}$ into value $\mathbf{V}=\mathbf{W}_V \mathbf{F}_{\mathbf{L}}$, where $\mathbf{W}$ is the combination of a $1\times 1$ point-wise convolution and a $3\times 3$ depth-wise convolution. We then perform cross-attention and get the output feature $\tilde{\mathbf{F}}$:
\begin{equation}
    \tilde{\mathbf{F}}=\mathbf{F}+\text{SoftMax}\left(\mathbf{Q}\mathbf{K}^{T}/\sqrt{\tilde C}\right)\cdot \mathbf{V}.
\end{equation}

By doing so, RG-MCA introduces explicit guidance to fully exploit Retinex knowledge at the feature level and use cross attention mechanism to implicitly model the Retinex theory and refine the decomposed features, which helps to restore missing details and correct color distortion. 

Then we employ DFA for local feature aggregation. Apart from the $1\times 1$ convolution and $3\times 3$ depth-wise convolution used for information fusion, DFA also adopts GELU, termed $\text{GELU}(\bigcdot)$, to ensure the flexibility of aggregation~\cite{he2023camouflaged}. Thus, given $\tilde{\mathbf{F}}$ and $\mathbf{Z}$, where $\mathbf{Z}=\text{conca}(\mathbf{Z}_\mathbf{R},\mathbf{Z}_\mathbf{L})$, the output feature $\hat{\mathbf{F}}$ is
\begin{equation}
\begin{aligned}
   &\hat{\mathbf{F}}= \tilde{\mathbf{F}} + \text{GELU}(\mathbf{W}_1 \mathbf{F}') \odot \mathbf{W}_2 \mathbf{F}', \\
   & \mathbf{F}'= \text{Li}_1 (\mathbf{Z})\odot \text{Norm}(\tilde{\mathbf{F}})+\text{Li}_2 (\mathbf{Z}),
\end{aligned}
\end{equation}

\noindent\textbf{Optimization.} To facilitate the extraction of Retinex priors, the RPE module and RGformer are jointly trained by a reconstruction loss with $L_1$ norm $\|\bigcdot\|_1$:
\begin{equation}
    L_{Rec}=\|\mathbf{I}_{GT}-\mathbf{I}_{HQ}\|_1,
\end{equation}
where $\mathbf{I}_{HQ}$ is the enhanced result.
In addition, to ensure that the separated features within RG-MCA effectively capture reflectance and illumination knowledge, we provide an auxiliary decoder $D_a(\bigcdot)$ with the same structure as that in~\cite{locatello2020object}. $D_a(\bigcdot)$ takes $\tilde{\mathbf{F}}$ as input and outputs the reconstructed reflectance image $\mathbf{R}_{Re}$ and illumination map $\mathbf{L}_{Re}$. For efficiency, we only apply $D_a(\bigcdot)$ for the first transformer block in encoder to get $\mathbf{R}_{Re}^I$ and $\mathbf{L}_{Re}^I$ and for the last transformer block in decoder to get $\mathbf{R}_{Re}^L$ and $\mathbf{L}_{Re}^L$. $D_a(\bigcdot)$ is supervised by a Retinex loss $L_{R}$:
\begin{equation}
\hspace{-2mm}
\label{Eq:RetLoss}
\begin{aligned}
    L_{R}=\|\mathbf{R}_{LQ} -\mathbf{R}_{Re}^I\|_1+\|\mathbf{L}_{LQ} -\mathbf{L}_{Re}^I\|_1+ \|\mathbf{R}_{GT} -\mathbf{R}_{Re}^L\|_1+\|\mathbf{L}_{GT} -\mathbf{L}_{Re}^L\|_1,
\end{aligned}
\end{equation}
\cref{Eq:RetLoss} serves to maintain crucial Retinex information throughout the network.
Hence, the integration of \cref{Eq:RetLoss} not only promotes the assimilation of Retinex theory by the split features but also amplifies the overall restoration capacity.


In Phase \uppercase\expandafter{\romannumeral1}, the final loss $L_{P1}$ is defined as follows: 
\begin{equation}
    L_{P1} = L_{Rec}+\lambda_1 L_{R},
\end{equation}
where $\lambda_1$ is a hyperparameter and $\lambda_1=1$. 

\subsection{Retinex-based Latent Diffusion Model} \vspace{-0.5mm}
In Phase \uppercase\expandafter{\romannumeral2}, we train the RLDM to predict Retinex priors from the low-quality input, which are expected to be consistent with that extracted by RPE from the ground-truth image. Unlike conventional LDMs trained on the RGB domain, we introduce two RLDMs with a Siamese structure and train them on distinct domains: the reflectance domain and the illumination domain. This approach, grounded in Retinex theory, equips our RLDM to generate a more generative reflectance prior $\hat{\mathbf{Z}}_{\mathbf{R}}$ to enhance image details, and a more harmonized illumination prior $\hat{\mathbf{Z}}_{\mathbf{L}}$ for color correction. Note that RLDM is constructed upon the conditional denoising diffusion probabilistic models, with both a forward diffusion process and a reverse denoising process.
To simplify, we provide a detailed derivation for $\hat{\mathbf{Z}}_{\mathbf{R}}$ herein, while that of $\hat{\mathbf{Z}}_{\mathbf{L}}$ can be found in the appendix.

\noindent\textbf{Diffusion process.}
In the diffusion process, we first use the pretrained RPE to extract the reflectance prior $\mathbf{Z}_{\mathbf{R}}$, which is treated as the starting point of the forward Markov process, \textit{i.e.}, $\mathbf{Z}_{\mathbf{R}}=\mathbf{Z}_{\mathbf{R}}^0$. We then gradually add Gaussian noise to $\mathbf{Z}_{\mathbf{R}}$ by $T$ iterations and each iteration can be defined as: 
\begin{equation}\label{eq:diff1}
    q\left(\mathbf{Z}_{\mathbf{R}}^t |\mathbf{Z}_{\mathbf{R}}^{t-1} \right)=\mathcal{N}\left(\mathbf{Z}_{\mathbf{R}}^t;\sqrt{1-\beta^t}\mathbf{Z}_{\mathbf{R}}^{t-1} , \beta^t \mathbf{I} \right),
\end{equation}
where $t=1,\cdots,T$. $\mathbf{Z}_{\mathbf{R}}^t$ denotes the noisy prior at time step $t$, $\beta^t$ is the predefined factor that controls the noise variance, and $\mathcal{N}$ is the Gaussian distribution. Following~\cite{kingma2013auto}, \cref{eq:diff1} can be simplified as follows:
\begin{equation}
  q\left(\mathbf{Z}_{\mathbf{R}}^t |\mathbf{Z}_{\mathbf{R}}^0 \right)=\mathcal{N}\left(\mathbf{Z}_{\mathbf{R}}^t;\sqrt{\bar{\alpha}^t}\mathbf{Z}_{\mathbf{R}}^0 , (1-\bar{\alpha}^t) \mathbf{I} \right),
\end{equation}
where $\alpha^t = 1-\beta^t$ and $\bar{{\alpha}}^t=\prod_{i=1}^t\alpha^i$.

\noindent\textbf{Reverse process.}
In the reverse process, RLDM aims to extract the reflectance prior from pure Gaussian noise. Thus, RLDM samples a Gaussian random noise map $\mathbf{Z}_{\mathbf{R}}^T$ and then gradually denoise it to run backward from $\mathbf{Z}_{\mathbf{R}}^T$ to $\mathbf{Z}_{\mathbf{R}}^0$:
\begin{equation}
    p\!\left(\mathbf{Z}_{\mathbf{R}}^{t-1} |\mathbf{Z}_{\mathbf{R}}^{t}, \mathbf{Z}_{\mathbf{R}}^0\!\right)\!= \!\mathcal{N}\!\left(\mathbf{Z}_{\mathbf{R}}^{t-1};\boldsymbol{\mu}^t(\mathbf{Z}_{\mathbf{R}}^{t}, \mathbf{Z}_{\mathbf{R}}^0), (\boldsymbol{\sigma}^t)^2\mathbf{I}\!\right),
\end{equation}
where mean $\boldsymbol{\mu}^t(\mathbf{Z}_{\mathbf{R}}^{t}, \mathbf{Z}_{\mathbf{R}}^0) = \frac{1}{\sqrt{\alpha^t}}(\mathbf{Z}_{\mathbf{R}}^{t} - \frac{1-\alpha^t}{\sqrt{1-\bar{\alpha}^t}}\boldsymbol{\epsilon})$ and variance $(\boldsymbol{\sigma}^t)^2=\frac{1-\bar{\alpha}^{t-1}}{1-\bar{\alpha}^t}\beta^t$. $\boldsymbol{\epsilon}$ denotes the noise in $\mathbf{Z}_{\mathbf{R}}^{t}$ and is the only uncertain variable. Following previous practice~\cite{xia2023diffir}, we employ a denoising network $\boldsymbol{\epsilon}_{\theta}(\bigcdot)$ to estimate $\theta$. To operate in the latent space, we further introduce another RPE module $\widetilde{\text{RPE}}(\bigcdot)$ to extract the conditional reflectance vector $\mathbf{V}_{\mathbf{R}}\in \mathbb{R}^{3C'}$ from the reflectance image $\mathbf{R}_{LQ}$ of the LQ image, \textit{i.e.}, $\mathbf{V}_{\mathbf{R}}=\widetilde{\text{RPE}}(\text{down}(\mathbf{R}_{LQ}))$. Therefore, the denoising network can be represented by $\boldsymbol{\epsilon}_{\theta}\left(\mathbf{Z}_{\mathbf{R}}^{t}, \mathbf{V}_{\mathbf{R}}, t\right)$. By setting the variance to $1-\alpha^t$, we get
\begin{equation} \label{eq:diff2}
\setlength{\abovedisplayskip}{0pt}
\setlength{\belowdisplayskip}{0pt}
\small
    \mathbf{Z}_{\mathbf{R}}^{t-1}\!=\!\frac{1}{\sqrt{\alpha^t}}(\mathbf{Z}_{\mathbf{R}}^{t} \!-\! \frac{1-\alpha^t}{\sqrt{1-\bar{\alpha}^t}}\boldsymbol{\epsilon}_{\theta}(\mathbf{Z}_{\mathbf{R}}^{t}, \mathbf{V}_{\mathbf{R}}, t))\!+\!\sqrt{1-\alpha^t}\boldsymbol{\epsilon}^t,
\end{equation}
where $\boldsymbol{\epsilon}^t \sim \mathcal{N}(0,\mathbf{I})$. By using~\cref{eq:diff2} for $T$ iterations, we can get the predicted prior $\hat{\mathbf{Z}}_{\mathbf{R}}$ and use it to guide RGformer for image restoration. Because the size of the predicted prior $\hat{\mathbf{Z}}_{\mathbf{R}}\in \mathbb{R}^{3C'}$ is much smaller than the original reflectance image $\mathbf{R}_{LQ} \in \mathbb{R}^{H\times W\times C}$, RLDM needs much less iterations than those image-level diffusion models~\cite{yi2023diff}. Thus, we run the complete $T$ iterations for the prior generation rather than randomly selecting one time step.

\noindent\textbf{Optimization.} Given the predicted priors $\hat{\mathbf{Z}}_{\mathbf{R}}$ and $\hat{\mathbf{Z}}_{\mathbf{L}}$, generated by two Siamese RLDMs with specific weights, we propose the diffusion loss to supervise them:
\begin{equation}
    L_{Dif}=\|\mathbf{Z}_{\mathbf{R}}-\hat{\mathbf{Z}}_{\mathbf{R}}\|_1+\|\mathbf{Z}_{\mathbf{L}}-\hat{\mathbf{Z}}_{\mathbf{L}}\|_1.
\end{equation}

For restoration quality, we propose joint training RPE, RGformer, and RLDM. Thus, the loss in Phase \uppercase\expandafter{\romannumeral2} is formulated as follows:
\begin{equation}
    L_{P2}= L_{Dif}+ \lambda_2 L_{Rec}+ \lambda_3 L_{R},
\end{equation}
where $\lambda_2$ and $\lambda_3$ are two hyper-parameters and are set as 1 in this paper.
\vspace{-3mm}
\subsection{Inference}\vspace{-1mm}
In the inference phase, given the LQ input $\mathbf{I}_{LQ}$, Reti-Diff first uses $\widetilde{\text{RPE}}$ to extract the conditional vectors $\mathbf{V}_{\mathbf{R}}$ and $\mathbf{V}_{\mathbf{L}}$, and then generates predicted Retinex priors $\hat{\mathbf{Z}}_{\mathbf{R}}$ and $\hat{\mathbf{Z}}_{\mathbf{L}}$ with two RLDMs. Under the guidance of the Retinex priors, RGformer generates the restored HQ image $\mathbf{I}_{HQ}$. Benefiting from our Retinex-based diffusion framework, $\mathbf{I}_{HQ}$ enjoys richer texture details and more harmonized illumination, thereby further facilitating downstream tasks.

\begin{table*}[t]
\centering
\setlength{\abovecaptionskip}{0cm}
\resizebox{\columnwidth}{!}{
\setlength{\tabcolsep}{0.6mm}
\begin{tabular}{l|c|cccc|cccc|cccc|cccc}
\toprule
                          &                       & \multicolumn{4}{c|}{\textit{LOL-v1}}& \multicolumn{4}{c|}{\textit{LOL-v2-real}}& \multicolumn{4}{c|}{\textit{LOL-v2-synthetic}}& \multicolumn{4}{c}{\textit{SID}}\\ \cline{3-18}
\multirow{-2}{*}{Methods} & \multirow{-2}{*}{Sources} & \cellcolor{gray!40}PSNR~$\uparrow$  & \cellcolor{gray!40}SSIM~$\uparrow$  & \cellcolor{gray!40}FID~$\downarrow$ & \cellcolor{gray!40}BIQE~$\downarrow$ & \cellcolor{gray!40}PSNR~$\uparrow$  & \cellcolor{gray!40}SSIM~$\uparrow$  & \cellcolor{gray!40}FID~$\downarrow$ & \cellcolor{gray!40}BIQE~$\downarrow$ & \cellcolor{gray!40}PSNR~$\uparrow$  & \cellcolor{gray!40}SSIM~$\uparrow$  & \cellcolor{gray!40}FID~$\downarrow$ & \cellcolor{gray!40}BIQE~$\downarrow$ & \cellcolor{gray!40}PSNR~$\uparrow$  & \cellcolor{gray!40}SSIM~$\uparrow$  & \cellcolor{gray!40}FID~$\downarrow$ & \cellcolor{gray!40}BIQE~$\downarrow$            \\ \midrule
MIRNet~\cite{zamir2020learning}                    & ECCV20                & 24.14                                 & 0.835                                  & 71.16                                 & 47.75                                 & 20.02                                 & 0.820                                 & 82.25                                 & 41.18                                 & 21.94                                 & 0.876                                 & 40.18                                 & 36.29                                 & 20.84                                 & 0.605                                 & 81.37                                 & 40.63                                 \\
EnGAN~\cite{jiang2021enlightengan}                     & TIP21                 & 17.48                                 & 0.656                                  & 153.98                                & 35.82                                 & 18.23                                 & 0.617                                 & 173.28                                & 51.06                                 & 16.57                                 & 0.734                                 & 93.66                                 & 45.59                                 & 17.23                                 & 0.543                                 & 77.52                                 & 33.47                                 \\
RUAS~\cite{liu2021retinex}                      & CVPR21                & 18.23                                 & 0.723                                  & 127.60                                & 45.17                                 & 18.27                                 & 0.723                                 & 151.62                                & 34.73                                 & 16.55                                 & 0.652                                 & 91.60                                 & 46.38                                 & 18.44                                 & 0.581                                 & 72.18                                 & 45.02                                 \\
IPT~\cite{chen2021pre}                       & CVPR21                & 16.27                                 & 0.504                                 & 158.83                                & 29.35                                 & 19.80                                 & 0.813                                 & 97.24                                 & 31.17                                 & 18.30                                 & 0.811                                 & 76.79                                 & 42.15                                 & 20.53                                 & 0.618                                 & 70.58                                 & 36.71                                 \\
URetinex~\cite{wu2022uretinex}             & CVPR22                & 21.33                                 & 0.835                                 & 85.59                                 & 30.37                                 & 20.44                                 & 0.806                                 & 76.74                                 & 28.85                                 & 24.73                                 & 0.897                                 & 33.25                                 & 33.46                                 & 22.09                                 & 0.633                                 & 71.58                                 & 38.44                                 \\
UFormer~\cite{wang2022uformer}                   & CVPR22                & 16.36                                 & 0.771                                 & 166.69                                & 41.06                                 & 18.82                                 & 0.771                                 & 164.41                                & 40.36                                 & 19.66                                 & 0.871                                 & 58.69                                 & 39.75                                 & 18.54                                 & 0.577                                 & 100.14                                & 42.13                                 \\
Restormer~\cite{zamir2022restormer}                 & CVPR22                & 22.43                                 & 0.823                                 & 78.75                                 & 33.18                                 & 19.94                                 & 0.827                                 & 114.35                                & 37.27                                 & 21.41                                 & 0.830                                 & 46.89                                 & 35.06                                 & 22.27                                 & 0.649                                 & 75.47                                 & 32.49                                 \\
SNR-Net~\cite{xu2022snr}                   & CVPR22                & 24.61                                 & 0.842                                 & 66.47                                 & 28.73                                 & 21.48                                 & 0.849                                 & 68.56                                 & 28.83                                 & 24.14                                 & 0.928                                 & 30.52                                 & 33.47                                 & 22.87                                 & 0.625                                 & 74.78                                 & 30.08                                 \\
SMG~\cite{xu2023low}                       & CVPR23                & 24.82                                 & 0.838                                 & 69.47                                 & 30.15                                 & 22.62                                 & 0.857                                 & 71.76                                 & 30.32                                 & 25.62                                 & 0.905                                 & 23.36                                 & 29.35                                 & 23.18                                 & 0.644                                 & 77.58                                 & 31.50                                 \\
PyDiff~\cite{zhou2023pyramid} & IJCAI23 &21.15 &{\color[HTML]{00B0F0} \textbf{0.857}} &{\color[HTML]{00B0F0} \textbf{49.47}} &21.13 &---&---&---&---&---&---&---&---&---&---&---&---   \\
Retformer~\cite{Cai2023}             & ICCV23                & 25.16                                 & 0.845                                 & 72.38                                 & 26.68                                 & {\color[HTML]{00B0F0} \textbf{22.80}} & 0.840                                 & 79.58                                 & 34.39                                 & {\color[HTML]{00B0F0} \textbf{25.67}} & 0.930                                 & 22.78                                 & 30.26                                 & 24.44                                 & 0.680                                 & 82.64                                 & 35.04                                 \\
Diff-Retinex~\cite{yi2023diff}              & ICCV23                & 21.98                                 & 0.852                                 & 51.33 & {\color[HTML]{00B0F0} \textbf{19.62}} & 20.17                                 & 0.826                                 & {\color[HTML]{00B0F0} \textbf{46.67}} & {\color[HTML]{00B0F0} \textbf{24.18}} & 24.30                                 & 0.921                                 & 28.74                                 & 26.35                                 & 23.62                                 & 0.665                                 & {\color[HTML]{00B0F0} \textbf{58.93}} & 31.17                                 \\
MRQ~\cite{liu2023low}     & ICCV23                & {\color[HTML]{00B0F0} \textbf{25.24}} & 0.855 & 53.32                                 & 22.73                                 & 22.37                                 & 0.854                                 & 68.89                                 & 33.61                                 & 25.54                                 & 0.940                                 & 20.86 & {\color[HTML]{00B0F0} \textbf{25.09}} & 24.62                                 & 0.683                                 & 61.09                                 & {\color[HTML]{00B0F0} \textbf{27.81}} \\
IAGC~\cite{wang2023low}                      & ICCV23                & 24.53                                 & 0.842                                 & 59.73                                 & 25.50                                 & 22.20                                 & {\color[HTML]{FF0000} \textbf{0.863}} & 70.34                                 & 31.70                                 & 25.58                                 & {\color[HTML]{00B0F0} \textbf{0.941}} & 21.38                                 & 30.32                                 & {\color[HTML]{00B0F0} \textbf{24.80}}                                 & {\color[HTML]{00B0F0} \textbf{0.688}} & 63.72                                 & 29.53                                 \\
DiffIR~\cite{xia2023diffir}                    & ICCV23                & 23.15                                 & 0.828                                 & 70.13                                 & 26.38                                 & 21.15                                 & 0.816                                 & 72.33                                 & 29.15                                 & 24.76                                 & 0.921                                 & 28.87                                 & 27.74                                 & 23.17                                 & 0.640                                 & 78.80                                 & 30.56                                 \\
CUE~\cite{zheng2023empowering}                       & ICCV23                & 21.86                                 & 0.841                                 & 69.83                                 & 27.15                                 & 21.19                                 & 0.829                                 & 67.05                                 & 28.83                                 & 24.41                                 & 0.917                                 & 31.33                                 & 33.83                                 & 23.25                                 & 0.652                                 & 77.38                                 & 28.85                                 \\
GSAD~\cite{jinhui2023global} & NIPS23 & 23.23  &0.852  &51.64 &19.96  & 20.19 & 0.847 &46.77 &28.85 & 24.22 &0.927 &{\color[HTML]{00B0F0} \textbf{19.24}} &25.76 &---&---&---&---  \\
Reti-Diff (Ours)                      & ---                    & {\color[HTML]{FF0000} \textbf{25.35}} & {\color[HTML]{FF0000} \textbf{0.866}} & {\color[HTML]{FF0000} \textbf{49.14}} & {\color[HTML]{FF0000} \textbf{17.75}} & {\color[HTML]{FF0000} \textbf{22.97}} & {\color[HTML]{00B0F0} \textbf{0.858}} & {\color[HTML]{FF0000} \textbf{43.18}} & {\color[HTML]{FF0000} \textbf{23.66}} & {\color[HTML]{FF0000} \textbf{27.53}} & {\color[HTML]{FF0000} \textbf{0.951}} & {\color[HTML]{FF0000} \textbf{13.26}} & {\color[HTML]{FF0000} \textbf{15.77}} & {\color[HTML]{FF0000} \textbf{25.53}} & {\color[HTML]{FF0000} \textbf{0.692}} & {\color[HTML]{FF0000} \textbf{51.66}} & {\color[HTML]{FF0000} \textbf{25.58}} \\ \bottomrule
\end{tabular}}
\caption{Results on the low-light image enhancement task. 
The best two results are in {\color[HTML]{FF0000}\textbf{red}} and {\color[HTML]{00B0F0}\textbf{blue}} fonts, respectively.}\label{table:low-light}\vspace{-5.5mm}
\end{table*}
\begin{figure*}[t]
	\centering
	\setlength{\abovecaptionskip}{-0.2cm}
	\begin{center}
		\includegraphics[width=\linewidth]{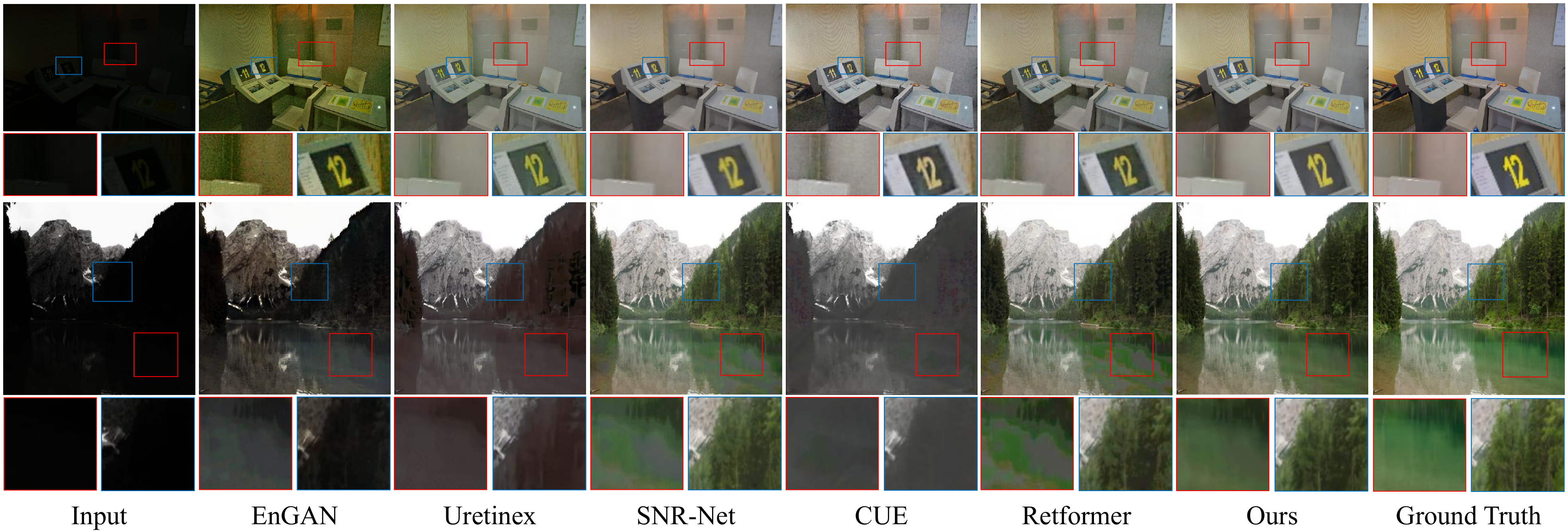}
	\end{center}
 \vspace{-1mm}
	\caption{Visual results on the low-light image enhancement task.}
	\label{fig:low-light}
	\vspace{-0.6cm}
\end{figure*}

\vspace{-1.5mm}
\section{Experiment}\vspace{-1.5mm}
\subsection{Experimental Setup}\vspace{-1mm}
Our Reti-Diff is implemented in PyTorch on four RTX3090TI GPUs and is optimized by Adam with momentum terms $(0.9,0.999)$. In phases \uppercase\expandafter{\romannumeral1} and \uppercase\expandafter{\romannumeral2}, we train the network for 300K iterations and the learning rate is initially set as $2\times 10^{-4}$ and gradually reduced to $1\times 10^{-6}$ with the cosine annealing~\cite{loshchilovstochastic}. Following~\cite{yi2023diff}, random rotation and flips are used for augmentation.
Reti-Diff mainly comprises RLDM and RGformer. For RLDM, the channel number $C'$ is set as 64.
The total time step $T$ is set to 4 and the hyperparameters $\beta^{1:T}$ linearly increase from $\beta^{1}=0.1$ to $\beta^{T}=0.99$.
RGformer adopts a 4-level cascade encoder-decoder structure. We set the number of transformer blocks, the attention heads, the channel number as $[3, 3, 3, 3]$, $[1, 2, 4, 8]$, $[64, 128, 256, 512]$ from level 1 to 4.

\begin{figure*}[t]
	\centering
	\setlength{\abovecaptionskip}{-0.2cm}
	\begin{center}
		\includegraphics[width=\linewidth]{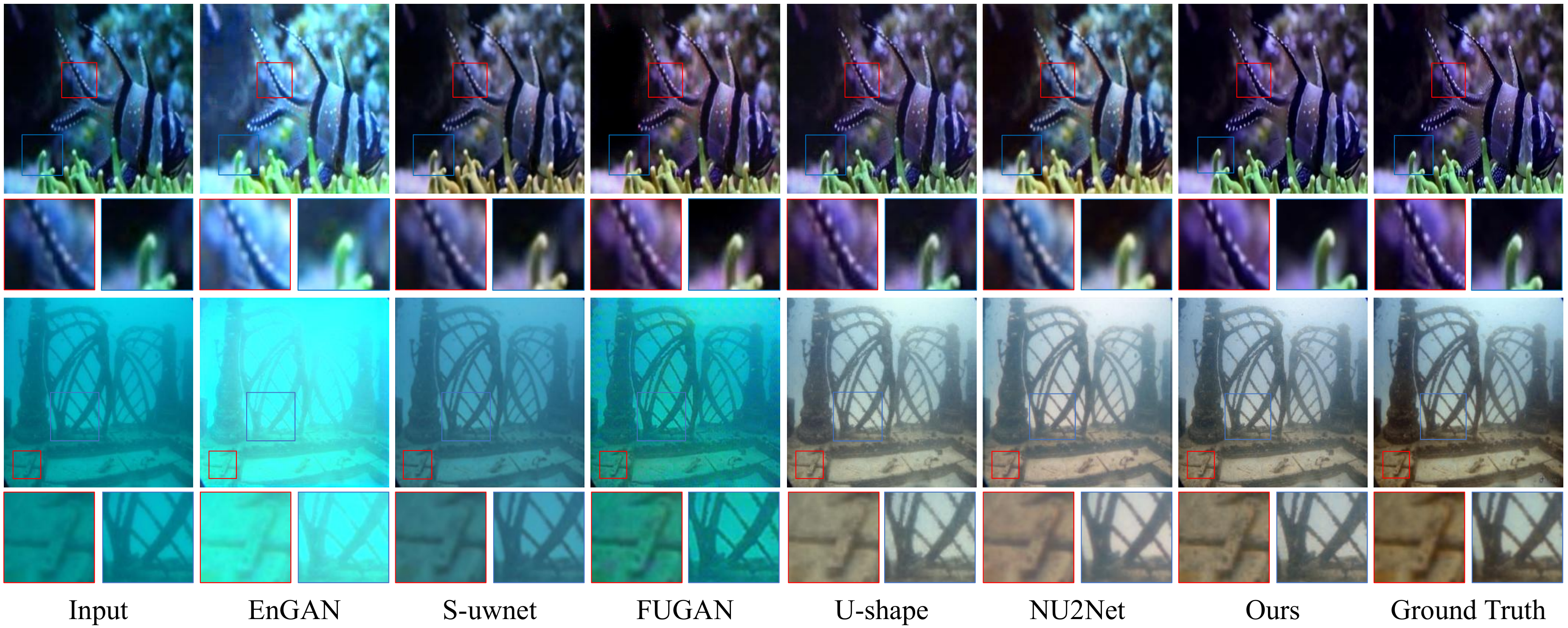}
	\end{center}
 \vspace{-1.3mm}
	\caption{Visual results on the underwater image enhancement task.}
	\label{fig:underwater}
	\vspace{-0.4cm}
\end{figure*}
\begin{figure*}[t]
	\centering
	\setlength{\abovecaptionskip}{-0.2cm}
	\begin{center}
		\includegraphics[width=\linewidth]{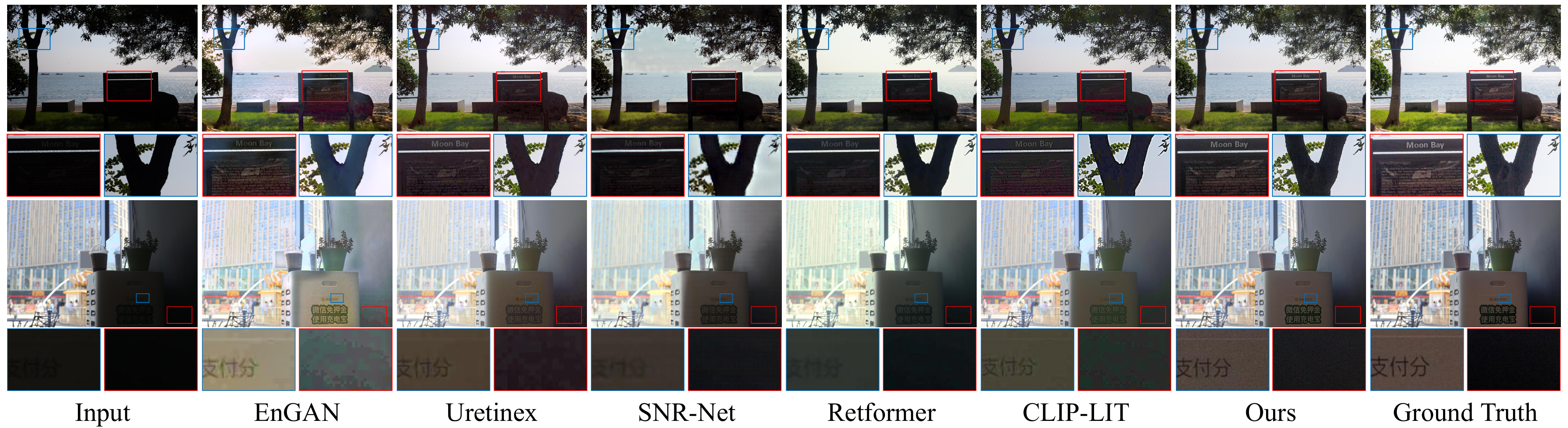}
	\end{center}
 \vspace{-1.3mm}
	\caption{Visual results on the backlit image enhancement task.}
	\label{fig:backlit}
	\vspace{-0.6cm}
\end{figure*}

\subsection{Comparative Evaluation}

\begin{table*}[t]
\begin{minipage}[c]{0.616\textwidth}
\centering
\setlength{\abovecaptionskip}{0cm}
\resizebox{\columnwidth}{!}{
\setlength{\tabcolsep}{0.6mm}
\begin{tabular}{l|c|cccc|cccc}
\toprule
 && \multicolumn{4}{c|}{\textit{UIEB}} & \multicolumn{4}{c}{\textit{LSUI}} \\ \cline{3-10}
\multirow{-2}{*}{Methods} &\multirow{-2}{*}{Sources} & \cellcolor{gray!40}PSNR~$\uparrow$ & \cellcolor{gray!40}SSIM~$\uparrow$& \cellcolor{gray!40}UCIQE~$\uparrow$& \cellcolor{gray!40}UIQM~$\uparrow$& \cellcolor{gray!40}PSNR~$\uparrow$& \cellcolor{gray!40}SSIM~$\uparrow$& \cellcolor{gray!40}UCIQE~$\uparrow$& \cellcolor{gray!40}UIQM~$\uparrow$\\ \midrule
FUGAN~\cite{islam2020fast}& IRAL20 & 17.41 & 0.842 & 0.527 & 2.614&22.16 &0.837&0.576 &2.667 \\
EnGAN~\cite{jiang2021enlightengan}& TIP21& 17.73                                 & 0.833                                 & 0.529                                 & 2.465                                 & 19.30                                 & 0.851                                 & 0.587                                 & 2.817                                 \\
Ucolor~\cite{li2021underwater}& TIP21& 20.78                                 & 0.868                                 & 0.537                                 & 3.049                                 & 22.91                                 & 0.886                                 & 0.594                                 & 2.735                                 \\
S-uwnet~\cite{naik2021shallow}& AAAI21&18.28                                 & 0.855                                 & 0.544                                 & 2.942                                 & 20.89                                 & 0.875                                 & 0.582                                 & 2.746                                 \\
PUIE~\cite{fu2022uncertainty}& ECCV22 & 21.38                                & 0.882                                 & 0.566                                 & {\color[HTML]{00B0F0} \textbf{3.021}} & 23.70                                 & 0.902                                 & 0.605                                 & 2.974                                 \\
U-shape~\cite{peng2023u}& TIP23 & 22.91                                 & {\color[HTML]{00B0F0} \textbf{0.905}} & 0.592                                 & 2.896                                 & 24.16                                 & {\color[HTML]{00B0F0} \textbf{0.917}} & 0.603                                 & 3.022                                 \\
PUGAN~\cite{cong2023pugan}& TIP23& {\color[HTML]{00B0F0} \textbf{23.05}} & 0.897                                 & 0.608                                 & 2.902                                 & 25.06                                 & 0.916                                 & {\color[HTML]{00B0F0} \textbf{0.629}} & 3.106                                 \\
ADP~\cite{zhou2023underwater}& IJCV23   & 22.90                                 & 0.892                                 & {\color[HTML]{00B0F0} \textbf{0.621}} & 3.005                                 & 24.28                                 & 0.913                                 & 0.626                                 & 3.075                                 \\
NU2Net~\cite{guo2023underwater}   & AAAI23 & 22.38                                 & 0.903                                 & 0.587                                 & 2.936                                 & {\color[HTML]{00B0F0} \textbf{25.07}} & 0.908                                 & 0.615                                 & {\color[HTML]{00B0F0} \textbf{3.112}} \\
Reti-Diff (Ours)              & ---        & {\color[HTML]{FF0000} \textbf{24.12}} & {\color[HTML]{FF0000} \textbf{0.910}} & {\color[HTML]{FF0000} \textbf{0.631}} & {\color[HTML]{FF0000} \textbf{3.088}} & {\color[HTML]{FF0000} \textbf{28.10}} & {\color[HTML]{FF0000} \textbf{0.929}} & {\color[HTML]{FF0000} \textbf{0.646}} & {\color[HTML]{FF0000} \textbf{3.208}} \\ \bottomrule
\end{tabular}}
\caption{Results on the underwater image enhancement task.}
		\label{table:Underwater} \vspace{-7.5mm}
\end{minipage}
\begin{minipage}[c]{0.38\textwidth}
\centering
\setlength{\abovecaptionskip}{0cm}
\resizebox{\columnwidth}{!}{
\setlength{\tabcolsep}{0.6mm}
\begin{tabular}{l|c|cccc}
\toprule
&  & \multicolumn{4}{c}{\textit{BAID}}                                                                                                                             \\ \cline{3-6}
\multirow{-2}{*}{Methods} & \multirow{-2}{*}{Sources} & \cellcolor{gray!40}PSNR~$\uparrow$ & \cellcolor{gray!40}SSIM~$\uparrow$ & \cellcolor{gray!40}LPIPS~$\downarrow$ & \cellcolor{gray!40}FID~$\downarrow$ \\ \midrule
EnGAN~\cite{jiang2021enlightengan}   & TIP21                       & 17.96                                 & 0.819                                 & 0.182                                 & 43.55                                 \\
RUAS~\cite{liu2021retinex}   & CVPR21                      & 18.92                                 & 0.813                                 & 0.262                                 & 40.07                                 \\
URetinex~\cite{wu2022uretinex}    & CVPR22                      & 19.08                                 & 0.845                                 & 0.206                                 & 42.26                                 \\
SNR-Net~\cite{xu2022snr}                   & CVPR22                      & 20.86                                 & 0.860                                 & 0.213                                 & 39.73                                 \\
Restormer~\cite{zamir2022restormer}                 & CVPR22                      & 21.07                                 & 0.832                                 & 0.192                                 & 41.17                                 \\
Retformer~\cite{Cai2023}             & ICCV23                      & 22.03                                 & {\color[HTML]{00B0F0} \textbf{0.862}} & 0.173                                 & 45.27                                 \\
CLIP-LIT~\cite{liang2023iterative}                  & ICCV23                      & 21.13                                 & 0.853                                 & {\color[HTML]{00B0F0} \textbf{0.159}} & {\color[HTML]{00B0F0} \textbf{37.30}} \\
Diff-Retinex~\cite{yi2023diff}              & ICCV23                      & {\color[HTML]{00B0F0} \textbf{22.07}} & 0.861                                 & 0.160                                 & 38.07                                 \\
DiffIR~\cite{xia2023diffir}                    & ICCV23                      & 21.10                                 & 0.835                                 & 0.175                                 & 40.35                                 \\
Reti-Diff (Ours)                      & ---                     & {\color[HTML]{FF0000} \textbf{23.19}} & {\color[HTML]{FF0000} \textbf{0.876}} & {\color[HTML]{FF0000} \textbf{0.147}} & {\color[HTML]{FF0000} \textbf{27.47}} \\ \bottomrule
\end{tabular}}
\caption{Results on the backlit image enhancement task.}
\label{table:backlit} \vspace{-7.5mm}
\end{minipage}
\end{table*}

\begin{table}[t]
\centering
\setlength{\abovecaptionskip}{0cm}
\resizebox{0.8\columnwidth}{!}{
\setlength{\tabcolsep}{1mm}
\begin{tabular}{l|c|cc|cc|cc|cc|cc}
\toprule
\multirow{2}{*}{Methods}                            & \multicolumn{1}{c|}{\multirow{2}{*}{Sources}} & \multicolumn{2}{c|}{\textit{DICM}} & \multicolumn{2}{c|}{\textit{LIME}} & \multicolumn{2}{c|}{\textit{MEF}} & \multicolumn{2}{c|}{\textit{NPE}} & \multicolumn{2}{c}{\textit{VV}} \\
& \multicolumn{1}{c|}{}                         &\cellcolor{gray!40}PI~$\downarrow$       & \cellcolor{gray!40}NIQE~$\downarrow$              & \cellcolor{gray!40}PI~$\downarrow$       & \cellcolor{gray!40}NIQE~$\downarrow$              & \cellcolor{gray!40}PI~$\downarrow$      & \cellcolor{gray!40}NIQE~$\downarrow$              & \cellcolor{gray!40}PI~$\downarrow$      & \cellcolor{gray!40}NIQE~$\downarrow$              & \cellcolor{gray!40}PI~$\downarrow$      & \cellcolor{gray!40}NIQE~$\downarrow$             \\ \midrule
EnGAN~\cite{jiang2021enlightengan}                                               & TIP21                                        &   4.173    & 4.064             &   3.669     & 4.593             &     4.015     & 4.705             &   3.226  & 3.993             & 3.386  & 4.047            \\
KinD++~\cite{zhang2021beyond}                                              & IJCV21                        &        3.835     & 3.898             &        3.785       & 4.908             &       4.016       & 4.557             &      3.179        & 3.915             &       3.773       & 3.822            \\
SNR-Net~\cite{xu2022snr}                                             & CVPR22                                       &   3.585   & 4.715             &  3.753     & 5.937             &   3.677   & 6.449             &   3.278    & 6.446             & 3.503     & 9.506            \\
DCC-Net~\cite{zhang2022deep}                                             & CVPR22                                       &      3.630         & 3.709             &   \color[HTML]{00B0F0}  \textbf{3.312}          &\color[HTML]{00B0F0} \textbf{4.425}             &  \color[HTML]{00B0F0} \textbf{3.424}           & 4.598             &      \color[HTML]{00B0F0}   \textbf{2.878}     &\color[HTML]{00B0F0} \textbf{3.706}             &       3.615       &\color[HTML]{00B0F0} \textbf{3.286}            \\
UHDFor~\cite{UHDFourICLR2023} &ICLR23& 3.684 &4.575 & 4.124 & 4.430 &  3.813 & 4.231 & 3.135 & 3.867 & 3.319  &4.330 \\
PairLIE~\cite{fu2023learning} & CVPR23                                       & 3.685 & 4.034             &  3.387   & 4.587             &  4.133  & 4.065             &  3.726  & 4.187             &\color[HTML]{FF0000} \textbf{3.334}  & 3.574            \\
GDP~\cite{fei2023generative} & CVPR23 & \color[HTML]{00B0F0} \textbf{3.552}  & 4.358&   4.115 & 4.891 &  3.694  & 4.609  &  3.097   & 4.032 & 3.431    & 4.683 \\
GSAD~\cite{jinhui2023global}  & NIPS23  &  ---     &\color[HTML]{00B0F0} \textbf{3.465}             &  ---   & 4.517             &    ---  &\color[HTML]{00B0F0} \textbf{3.815}             &  ---   & 3.806             &   ---   & 3.355            \\
Reti-Diff (Ours) & --- &\color[HTML]{FF0000} \textbf{2.351}     &\color[HTML]{FF0000} \textbf{3.255}    & \color[HTML]{FF0000} \textbf{2.837}     & \color[HTML]{FF0000} \textbf{3.693}    & \color[HTML]{FF0000} \textbf{3.308} &\color[HTML]{FF0000} \textbf{3.792}    & \color[HTML]{FF0000} \textbf{2.599}     &\color[HTML]{FF0000} \textbf{3.384}    &\color[HTML]{00B0F0}  \textbf{3.341}    &\color[HTML]{FF0000} \textbf{3.000}  \\ \bottomrule
\end{tabular}}
\caption{Results on the real-world illumination degradation image restoration task.}
\label{table:realIDIR} \vspace{-10mm}
\end{table}

\noindent\textbf{Low-light Image Enhancement.} We conduct a comprehensive evaluation on four datasets: \textit{LOL-v1}~\cite{wei2018deep}, \textit{LOL-v2-real}~\cite{yang2021sparse}, \textit{LOL-v2-syn}~\cite{yang2021sparse}, and \textit{SID}~\cite{chen2019seeing}. We adhere to the training manner outlined in~\cite{Cai2023}. Our assessment involves four metrics: PSNR, SSIM, FID~\cite{heusel2017gans}, and BIQE~\cite{moorthy2010two}.
Note that larger PSNR and SSIM scores, as well as smaller FID and BIQE scores, denote superior performance. We compare our approach against 17 cutting-edge enhancement techniques and report the results in~\cref{table:low-light}. 
As depicted in~\cref{table:low-light}, our method emerges as\begin{wraptable}[5]{r}{0.45\textwidth}
\vspace{-8.2mm}
\setlength{\abovecaptionskip}{0cm} 
	\setlength{\belowcaptionskip}{0cm}
		\centering
		\resizebox{0.45\columnwidth}{!}{
			\setlength{\tabcolsep}{1mm}
\begin{tabular}{l|cccc}
\toprule
         &\cellcolor{c2!50} Diff-Retinex~\cite{yi2023diff}&\cellcolor{c2!50} PyDiff~\cite{zhou2023pyramid} &\cellcolor{c2!50} GSAD~\cite{jinhui2023global}     &\cellcolor{c2!50} Ours   \\ \midrule
Parameter (M) & 56.88   & 97.89     & \textbf{17.17}     & 26.11  \\
MACs (G) & 396.32   & 459.69    & 1340.63  & \textbf{156.55} \\
FPS      & 4.25    & 3.63     & 2.33       & \textbf{12.27} \\ \bottomrule
\end{tabular}}
	\caption{Efficiency analysis in diffusion model-based methods. 
 }\label{table:efficiency}
	\vspace{-5mm}
\end{wraptable} the top performer across all datasets and significantly outperforms the second-best 
method (Diff-Retinex) by $13.2\%$. These results underscore the superiority of our Reti-Diff.~\cref{fig:low-light} presents qualitative results, showcasing our capacity to generate enhanced images with corrected illumination and enhanced texture, even in extremely challenging conditions. In contrast, existing methods struggle to achieve the same level of performance such as the boundaries of power lines, color harmonization of lakes, and detailed textures of wooded areas. Besides, we also compare the efficiency of the diffusion model-based methods. As presented in~\cref{table:efficiency}, despite having the second smallest parameter count, our Reti-Diff has the lowest MACs, highest FPS, and superior performance (see~\cref{table:low-light}). This efficiency can be attributed to our utilization of the diffusion model within a low-dimensional compact latent space.
For fairness, results from the compared methods are generated by their provided models under the same settings with no GT-mean strategy.


\noindent\textbf{Underwater Image Enhancement.}  We extend our evaluation to encompass two widely-used underwater image enhancement datasets: \textit{UIEB}~\cite{li2019underwater} and \textit{LSUI}~\cite{peng2023u}. In addition to PSNR and SSIM, we employ two metrics specifically tailored for underwater images, namely UCIQE~\cite{yang2015underwater} and UIQM~\cite{panetta2015human}, to assess the performance of the ten enhancement approaches. In all cases, higher values indicate superior performance. The results are presented in~\cref{table:Underwater}. As showcased in~\cref{table:Underwater}, our method achieved the highest overall performance and outperformed the second-best method (PUGAN) by $4.48\%$. A qualitative analysis is presented in~\cref{fig:underwater}, illustrating our method's ability to correct underwater color aberrations and highlight fine texture details.

\noindent\textbf{Backlit Image Enhancement.}
Following CLIP-LIT~\cite{liang2023iterative}, we select the \textit{BAID}~\cite{lv2022backlitnet} dataset for training the network with an image size of $256\times256$. In addition to PSNR and SSIM, our evaluation incorporates LPIPS~\cite{zhang2018unreasonable} and FID~\cite{heusel2017gans} as metrics for evaluation, where lower LPIPS and FID denote superior performance. The evaluation results are reported in~\cref{table:backlit}. As demonstrated in~\cref{table:backlit}, our method excels in all metrics and generally outperformed the second-best method (CLIP-LIT) by $6.03\%$. Furthermore, a visual comparison in~\cref{fig:backlit} provides additional evidence of our superiority in detail reconstruction and color correction. 

\noindent\textbf{Real-world Illumination Degradation Image Restoration.} We also explore the applicability of our method in real-world IDIR tasks. Following the practice of CIDNet~\cite{feng2024you}, we selected five commonly-used real-world datasets, \ie, \textit{DICM}~\cite{lee2013contrast}, \textit{LIME}~\cite{guo2016lime}, \textit{MEF}~\cite{wang2013naturalness}, \textit{NPE}~\cite{ma2015perceptual}, and \textit{VV}\footnote{\url{https://sites.google.com/site/vonikakis/datasets}}, which only have the low-quality images without paired high-quality ground-truth. Therefore, akin to~\cite{feng2024you}, we leverage models pretrained on the \textit{LOL-v2-syn} dataset for inference and select PI~\cite{blau20182018} and NIQE~\cite{mittal2012making} as evaluation metrics. In both metrics, lower scores indicate better results. As presented in~\cref{table:realIDIR}, our method achieves optimal results and surpasses the second-based method (DCC-Net~\cite{zhang2022deep}) by $13.39\%$. This verifies the generalizability of our Reti-Diff in addressing unknown degradation scenarios. Note that all the methods abandon the GT-mean strategy.

\begin{table*}[t]
\setlength{\abovecaptionskip}{0.05cm}
    \begin{subtable}{.543\textwidth}
\centering
\setlength{\abovecaptionskip}{0.05cm}
\resizebox{\columnwidth}{!}{
\setlength{\tabcolsep}{0.6mm}
\begin{tabular}{l|l|c|ccc|c|c}
\toprule
\multirow{2}{*}{Datasets} & \multirow{2}{*}{Metrics} &{RLDM}                & \multicolumn{3}{c|}{RGformer}                  & Train                  & ---     \\ \cline{3-8}
                         & & \cellcolor{c2!50} w/o RLDM  & \cellcolor{c2!50} w/o DFA & \cellcolor{c2!50} w/o RG-MCA & \cellcolor{c2!50} w/o $D_a(\bigcdot)$ & \cellcolor{c2!50} w/o joint &\cellcolor{c2!50}  Ours  \\ \midrule
\multirow{2}{*}{\textit{L-v2-r}} & PSNR                     & 21.25 & 22.26 & 21.73 & 22.58 & 22.83 & 22.97 \\
& SSIM                     & 0.822 & 0.840 & 0.840 & 0.847 & 0.853 & 0.858 \\ \midrule
\multirow{2}{*}{\textit{L-v2-s}} & PSNR                     & 25.38    & 26.49   & 25.92      & 26.80    & 27.18     & 27.53 \\
                                 & SSIM                     & 0.918    & 0.925   & 0.913      & 0.944    & 0.947     & 0.951 \\ \bottomrule
\end{tabular}}
\caption{Break down ablation. }\vspace{-4mm}
		\label{table:Ablation}
    \end{subtable}
    \begin{subtable}{.456\textwidth}
    \centering
\setlength{\abovecaptionskip}{0.05cm}
\resizebox{\columnwidth}{!}{
\setlength{\tabcolsep}{1mm}
\begin{tabular}{l|l|cccc}
\toprule
\multirow{2}{*}{Datasets}         & \multirow{2}{*}{Metrics} &\cellcolor{c2!50}  w/o Reti- &\cellcolor{c2!50} w/ Reflect- &\cellcolor{c2!50} w/ Illumina- &\cellcolor{c2!50} w/ Retinex   \\
                                 &                          &\cellcolor{c2!50} -nex prior       &\cellcolor{c2!50} -ance prior          &\cellcolor{c2!50} -tion prior           &\cellcolor{c2!50} prior (Ours) \\ \midrule
\multirow{2}{*}{\textit{L-v2-r}} & PSNR                     & 21.63       & 22.13          & 22.35           & 22.97        \\
                                 & SSIM                     & 0.830       & 0.842          & 0.839           & 0.858        \\ \midrule
\multirow{2}{*}{\textit{L-v2-s}} & PSNR                     & 26.25       & 26.62          & 27.02           & 27.53        \\
                                 & SSIM                     & 0.939       & 0.945          & 0.941           & 0.951       \\ \bottomrule
\end{tabular}}
\caption{Effect of our Retinex prior. }\vspace{-4mm}
		\label{table:Retinex}
    \end{subtable}
    \caption{Ablation study on the low-light image enhancement task. }\vspace{-6mm}
    \label{table:AblationStudy}
\end{table*}
\begin{figure}[t]
\begin{minipage}[c]{0.518\textwidth}
\setlength{\abovecaptionskip}{0cm} 
	\setlength{\belowcaptionskip}{0cm}
		\centering
		\resizebox{1\columnwidth}{!}{
			\setlength{\tabcolsep}{1mm}
           \begin{tabular}{l|l|cc|cc}
           \toprule
Datasets                         & Metrics &\cellcolor{c2!50}  Res~\cite{zamir2022restormer} &\cellcolor{c2!50}  Res+RLDM & \cellcolor{c2!50} Ret~\cite{Cai2023} &\cellcolor{c2!50}  Ret+RLDM \\ \midrule
\multirow{3}{*}{\textit{L-v2-s}} & PSNR    &  21.41     & 24.15          & 25.67     & 26.81          \\
                                 & SSIM    &  0.830     & 0.862          & 0.930     & 0.942          \\
                                 & Gain    &  --         & 8.33\%         & --         & 2.87\%         \\ \midrule
\multirow{3}{*}{\textit{L-v2-r}} & PSNR    & 19.94     & 21.56          & 22.80     & 23.16          \\
                                 & SSIM    & 0.827     & 0.837          & 0.840     & 0.849          \\
                                 &  Gain       & --         & 4.67\%         & --         & 1.33\%   \\ \bottomrule     
\end{tabular}}\vspace{0.5mm}		
	\captionof{table}{Generalization of our RLDM. ``Res'' and ``Ret'' are Restormer and Retformer.
	}\label{table:Generalization}
	\vspace{-5mm}
\end{minipage}
\begin{minipage}[c]{0.47\textwidth}
\centering
\setlength{\abovecaptionskip}{0.05cm}
    \begin{subfigure}{0.465\textwidth}
		\centering
		\includegraphics[width=\textwidth]{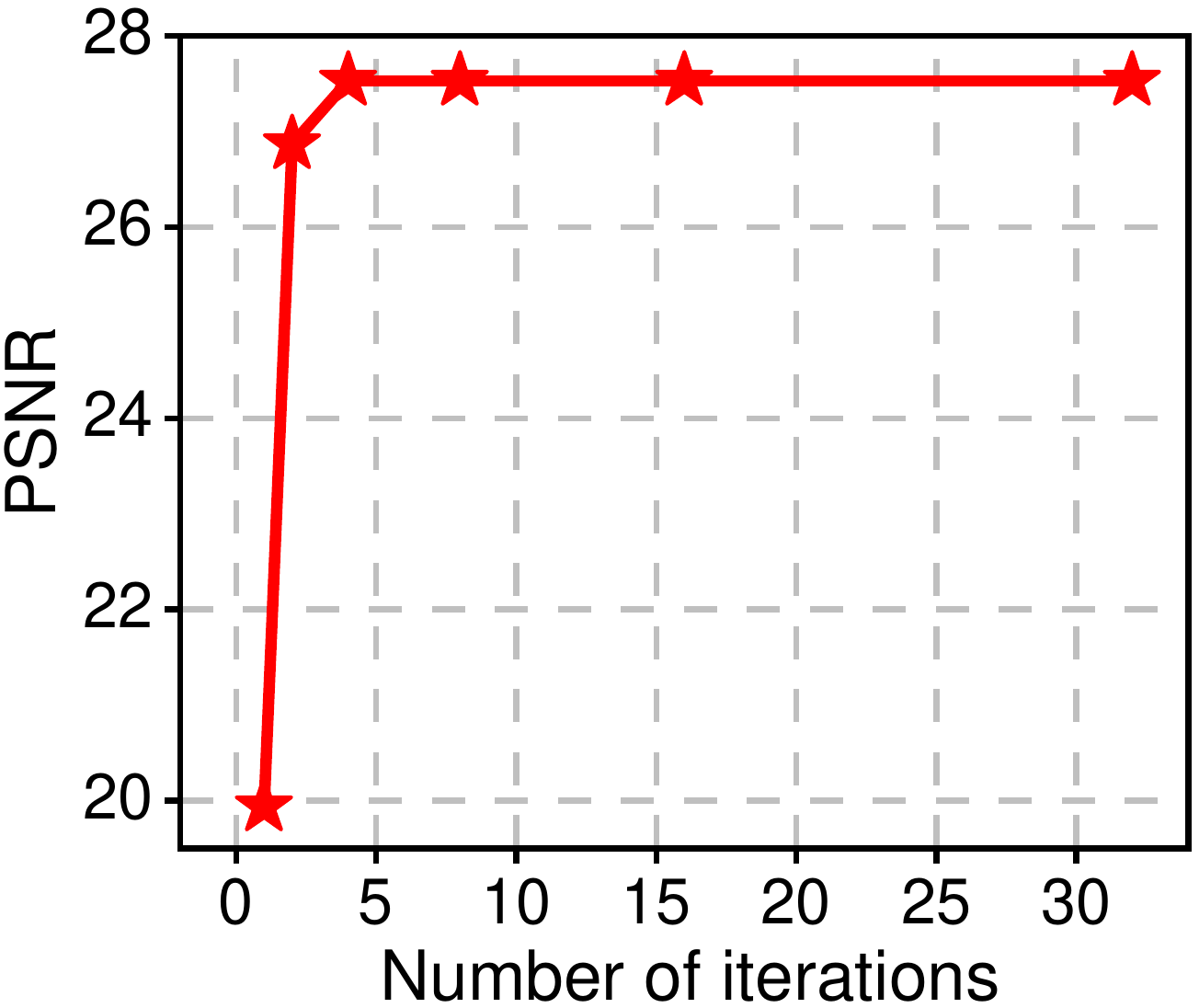}
	\end{subfigure}
	\hfill
	\begin{subfigure}{0.49\textwidth}  
		\centering 
		\includegraphics[width=\textwidth]{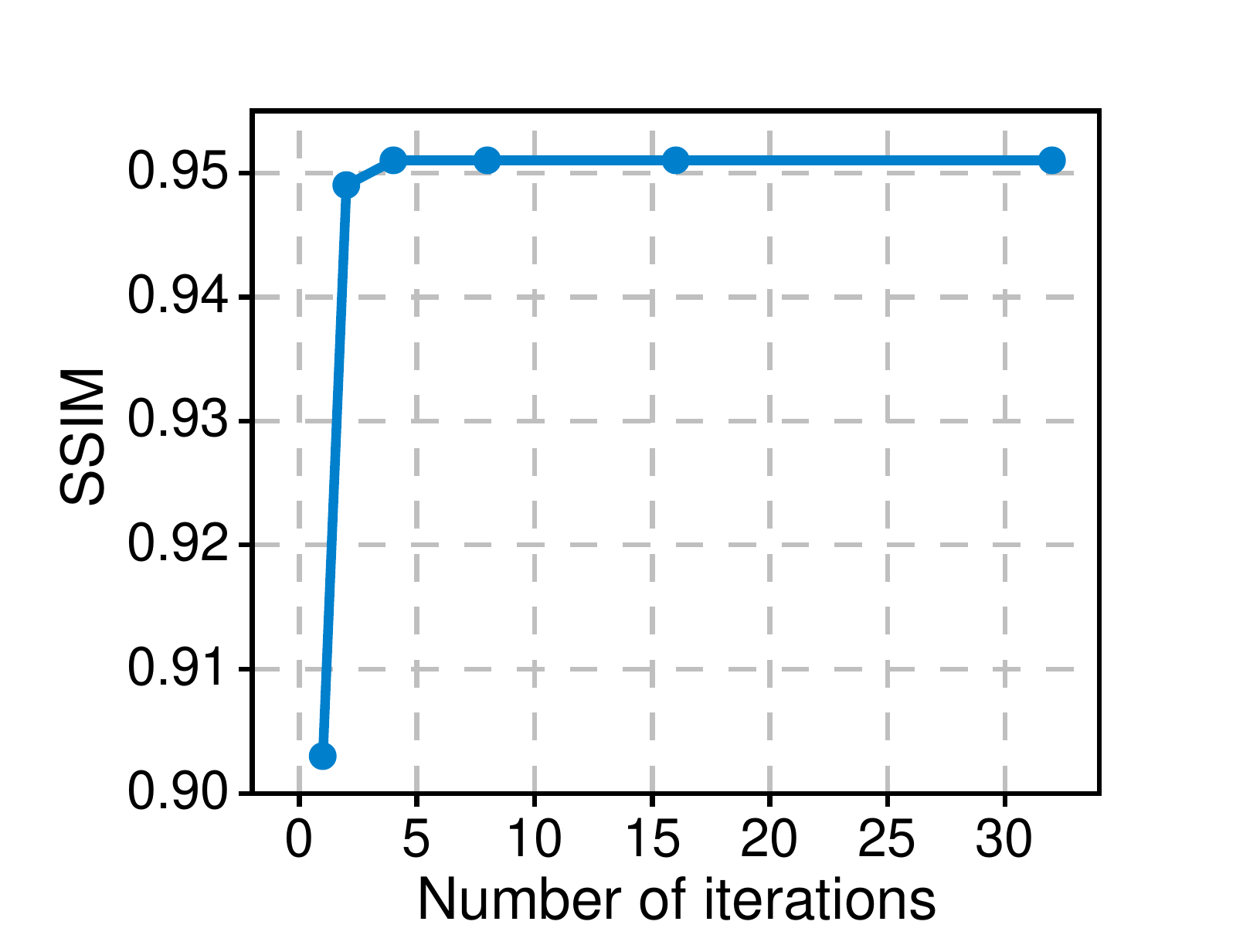}
	\end{subfigure}
	\caption{Ablation study of the number of iterations in RLDM on \textit{LOL-v2-syn}.}
	\label{fig:Ablation}
	\vspace{-5mm}
\end{minipage}
\end{figure}

\subsection{Ablation Study}

We conduct ablation studies on the low-light image enhancement task with the \textit{L-v2-r} and \textit{L-v2-s} datasets, which are short for \textit{LOL-v2-real} and \textit{LOL-v2-syn}.

\noindent\textbf{Effect of RLDM.}
As illustrated in~\cref{table:Ablation}, we ablate RLDM by directly removing RLDM or retraining RLDM in the RGB domain, \ie, w/o Retinex, rather than in the reflectance and illumination domain (RGformer is guided by one RGB prior instead of the Retinex priors in this time). The two modifications result in significant drops in performance. This outcome underscores the critical role of RLDM in enhancing the restoration process.
Furthermore, to assess the generalizability of RLDM, we conducted additional experiments by replacing our RGformer with two transformer-based frameworks, namely Res (Restormer~\cite{zamir2022restormer}) and Ret (Retformer~\cite{Cai2023}). Note that the training settings are kept consistent with our Reti-Diff. The results are presented in~\cref{table:Generalization}. \cref{table:Generalization} reveals that RLDM significantly improves the performance of both frameworks, where ``Gain'' is the average gain of PSNR and SSIM. This demonstrates that our RLDM serves as a plug-and-play module with strong generalization capabilities.

\noindent\textbf{Effect of RGformer.} We conduct an analysis to assess the impact of our RGformer, and the results are presented in~\cref{table:Ablation}. In this study, we systematically removed critical components, such as DFA, RG-MCA, and the auxiliary decoder $D_a(\bigcdot)$, from the model architecture. The outcomes of this ablation study clearly indicate that the performance deteriorates when these components are removed, highlighting their essential role in the system. Additionally, in~\cref{table:Ablation}, we conduct an evaluation to affirm the significance of joint training in our approach. This analysis reinforces the importance of the joint training process.

\noindent\textbf{Ablations on iteration number.} 
The number of iterations in the diffusion 
model plays a crucial role in determining the method's efficiency. To explore this, we conducted experiments with different iteration numbers for Reti-Diff, specifically $T$ values selected from the set $\{1,2,4,8,16,32\}$. We adjusted $\beta^t$ as defined in~\cref{eq:diff1} accordingly. The results in terms of PSNR for different iterations, as shown in~\cref{fig:Ablation}, illustrate that Reti-Diff exhibits rapid convergence and generates stable guidance priors with just $4$ iterations. This efficiency is attributed to our application of the diffusion model within the compact latent space. 


\begin{table*}[t]
\begin{minipage}[c]{0.356\textwidth}
\centering
\setlength{\abovecaptionskip}{0cm}
\resizebox{\columnwidth}{!}{
\setlength{\tabcolsep}{0.6mm}
\begin{tabular}{l|cccc|c}
\toprule
Methods &\multicolumn{1}{c}{\cellcolor{c2!50} \textit{L-v1}}           & \multicolumn{1}{c}{\cellcolor{c2!50} \textit{L-v2-r}}      & \multicolumn{1}{c}{\cellcolor{c2!50} \textit{L-v2-s}}       & \multicolumn{1}{c|}{\cellcolor{c2!50} \textit{SID}}              & \multicolumn{1}{c}{\cellcolor{c2!50} Mean}             \\ \midrule
KinD~\cite{zhang2019kindling} & 2.31 & 2.25 & 2.46 & 2.33 & 2.34  \\
EnGAN~\cite{jiang2021enlightengan}    & 2.63                                 & 1.69                                 & 2.23                                 & 1.24                                 & 1.95                                 \\
RUAS~\cite{liu2021retinex}& 3.57                                 & 3.06                                 & 3.01                                 & 2.23                                 & 2.97                                 \\
Restormer~\cite{zamir2022restormer}& 3.26                                 & 3.32                                 & 3.41                                 & 2.53                                 & 3.13                                 \\
Uretinex~\cite{wu2022uretinex}   & {\color[HTML]{00B0F0} \textbf{3.82}} & 3.98                                 & 3.70                                 & 3.28                                 & 3.70                                 \\
SNR-Net~\cite{xu2022snr}             & 3.76                                 & {\color[HTML]{00B0F0} \textbf{4.12}} & 3.58                                 & {\color[HTML]{00B0F0} \textbf{3.42}} & {\color[HTML]{00B0F0} \textbf{3.72}} \\
CUE~\cite{zheng2023empowering}& 3.62                                 & 3.81                                 & 3.28                                 & 3.09                                 & 3.45                                 \\
Retformer~\cite{Cai2023}   & 3.35                                 & 4.02                                 & {\color[HTML]{00B0F0} \textbf{3.71}} & 3.35                                 & 3.61                                 \\
Ours & {\color[HTML]{FF0000} \textbf{4.05}} & {\color[HTML]{FF0000} \textbf{4.33}} & {\color[HTML]{FF0000} \textbf{3.92}} & {\color[HTML]{FF0000} \textbf{3.75}} & {\color[HTML]{FF0000} \textbf{4.01}} \\ \bottomrule
\end{tabular}}
\caption{ User study. }
		\label{table:UserStudy} \vspace{-6mm}
\end{minipage}
\begin{minipage}[c]{0.63\textwidth}
\centering
\setlength{\abovecaptionskip}{0cm}
\resizebox{\columnwidth}{!}{
\setlength{\tabcolsep}{0.6mm}
\begin{tabular}{l|cccccccccccc|c}
\toprule
Methods (AP)& \cellcolor{c2!50}Bicycle  & \cellcolor{c2!50}Boat & \cellcolor{c2!50}Bottle  & \cellcolor{c2!50}Bus & \cellcolor{c2!50}Car & \cellcolor{c2!50}Cat  & \cellcolor{c2!50}Chair &\cellcolor{c2!50} Cup & \cellcolor{c2!50}Dog & \cellcolor{c2!50} Motor & \cellcolor{c2!50}People & \cellcolor{c2!50}Table  & \cellcolor{c2!50}Mean \\ \midrule
Baseline & 74.7                                 & 64.9                                 & 70.7                                 & 84.2                                 & 79.7                                 & 47.3                                 & 58.6                                 & 67.1                                 & 64.1                                 & 66.2                                 & 73.9                                 & 45.7                                 & 66.4                                 \\
RetinexNet~\cite{wei2018deep}& 72.8                                 & 66.4                                 & 67.3                                 & 87.5                                 & 80.6                                 & 52.8                                 & 60.0                                 & 67.8                                 & 68.5                                 & {\color[HTML]{00B0F0} \textbf{69.3}} & 71.3                                 & 46.2                                 & 67.5                                 \\
KinD~\cite{zhang2019kindling} & 73.2                                 & 67.1                                 & 64.6                                 & 86.8                                 & 79.5                                 & 58.7                                 & 63.4                                 & 67.5                                 & 67.4                                 & 62.3                                 & {\color[HTML]{00B0F0} \textbf{75.5}} & 51.4                                 & 68.1                                 \\
MIRNet~\cite{zamir2020learning}& 74.9                                 & 69.7                                 & 68.3                                 & 89.7                                 & 77.6                                 & 57.8                                 & 56.9                                 & 66.4                                 & 69.7                                 & 64.6                                 & 74.6                                 & 53.4                                 & 68.6                                 \\
RUAS~\cite{liu2021retinex} & 75.7                                 & 71.2                                 & 73.5                                 & 90.7                                 & 80.1                                 & 59.3                                 & {\color[HTML]{00B0F0} \textbf{67.0}} & 66.3                                 & 68.3                                 & 66.9                                 & 72.6                                 & 50.6                                 & 70.2                                 \\
Restormer~\cite{zamir2022restormer}       & 77.0                                 & 71.0                                 & 68.8                                 & {\color[HTML]{00B0F0} \textbf{91.6}} & 77.1                                 & 62.5                                 & 57.3                                 & {\color[HTML]{00B0F0} \textbf{68.0}} & 69.6                                 & 69.2                                 & 74.6                                 & 49.7                                 & 69.7                                 \\
SCI~\cite{ma2022toward} & 73.4                                 & 68.0                                 & 69.5                                 & 86.2                                 & 74.5                                 & 63.1                                 & 59.5                                 & 61.0                                 & 67.3                                 & 63.9                                 & 73.2                                 & 47.3                                 & 67.2                                 \\
SNR-Net~\cite{xu2022snr}         & {\color[HTML]{00B0F0} \textbf{78.3}} & 74.2                                 & {\color[HTML]{00B0F0} \textbf{74.5}} & 89.6                                 & {\color[HTML]{00B0F0} \textbf{82.7}} & {\color[HTML]{00B0F0} \textbf{66.8}} & 66.3                                 & 62.5                                 & 74.7                                 & 63.1                                 & 73.3                                 & {\color[HTML]{00B0F0} \textbf{57.2}} & 71.9                                 \\
Retformer~\cite{Cai2023}   & 78.1                                 & {\color[HTML]{00B0F0} \textbf{74.5}} & 74.2                                 & 91.2                                 & 82.2                                 & 65.0                                 & 63.3                                 & 67.0                                 & {\color[HTML]{00B0F0} \textbf{75.4}} & 68.6                                 & 75.3                                 & 55.6                                 & {\color[HTML]{00B0F0} \textbf{72.5}} \\
Ours & {\color[HTML]{FF0000} \textbf{82.0}} & {\color[HTML]{FF0000} \textbf{77.9}} & {\color[HTML]{FF0000} \textbf{76.4}} & {\color[HTML]{FF0000} \textbf{92.2}} & {\color[HTML]{FF0000} \textbf{83.3}} & {\color[HTML]{FF0000} \textbf{69.6}} & {\color[HTML]{FF0000} \textbf{67.4}} & {\color[HTML]{FF0000} \textbf{74.4}} & {\color[HTML]{FF0000} \textbf{75.5}} & {\color[HTML]{FF0000} \textbf{74.3}} & {\color[HTML]{FF0000} \textbf{78.3}} & {\color[HTML]{FF0000} \textbf{57.9}} & {\color[HTML]{FF0000} \textbf{75.8}}\\ \bottomrule
\end{tabular}}
\caption{Low-light image detection on \textit{ExDark}~\cite{loh2019getting}. }
		\label{table:Detection} \vspace{-6mm}
\end{minipage}
\end{table*}
\begin{figure*}[t]
	\centering
	\setlength{\abovecaptionskip}{-0.2cm}
	\begin{center}
		\includegraphics[width=\linewidth]{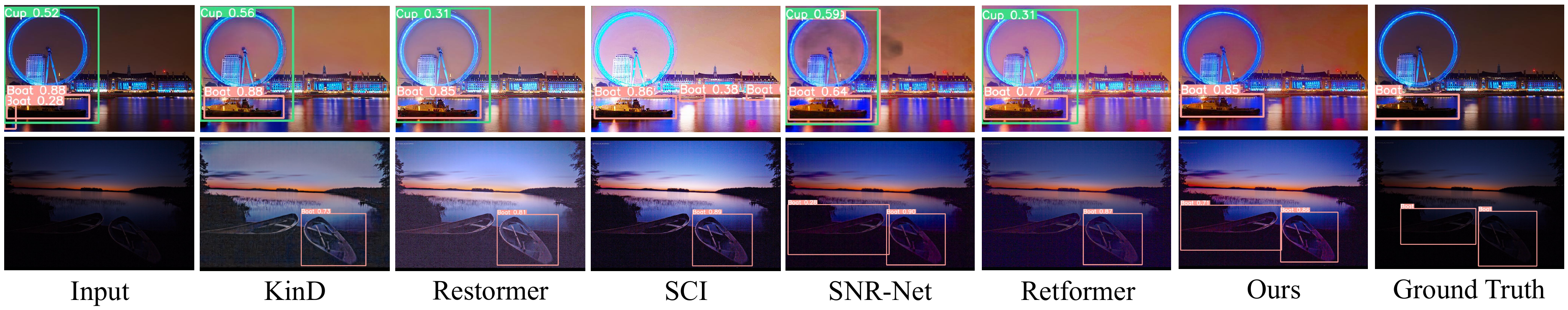}
	\end{center}
 \vspace{-1.3mm}
	\caption{Results on the low-light object detection task.}
	\label{fig:detection}
	\vspace{-0.6cm}
\end{figure*}

\begin{table*}[t]
\begin{minipage}[c]{0.562\textwidth}
\centering
\setlength{\abovecaptionskip}{0cm}
\resizebox{\columnwidth}{!}{
\setlength{\tabcolsep}{0.6mm}
\begin{tabular}{l|cccccccccc|c}
\toprule
Methods (IoU)        & \cellcolor{c2!50}Bicycle                               & \cellcolor{c2!50}Boat                                  & \cellcolor{c2!50}Bottle                                & \cellcolor{c2!50}Bus                                   & \cellcolor{c2!50}Car                                   & \cellcolor{c2!50}Cat                                   & \cellcolor{c2!50}Chair                                 & \cellcolor{c2!50}Dog                                   & \cellcolor{c2!50}Horse                                 & \cellcolor{c2!50}People                                & \cellcolor{c2!50}Mean                                  \\ \midrule
Baseline& 43.5                                 & 36.3                                 & 48.6                                 & 70.5                                 & 67.3                                 & 46.6                                 & 11.2                                 & 42.4                                 & 56.7                                 & 57.8                                 & 48.1                                   \\
RetinexNet~\cite{wei2018deep}      & 48.6                                 & 41.7                                 & 51.7                                 & 77.6                                 & 68.3                                 & 52.7                                 & 15.8                                 & 46.3                                 & 60.2                                 & 62.3                                 & 52.5                                 \\
KinD~\cite{zhang2019kindling}            & 51.3                                 & 40.2                                 & 53.2                                 & 76.8                                 & 69.4                                 & 50.8                                 & 14.6                                 & 47.3                                 & 60.3                                 & 60.9                                 & 52.5                                 \\
MIRNet~\cite{zamir2020learning}          & 50.3                                 & 42.9                                 & 47.4                                 & 73.6                                 & 62.7                                 & 50.4                                 & 15.8                                 & 46.3                                 & 61.0                                 & 63.3                                 & 51.4                                 \\
RUAS~\cite{liu2021retinex}           & 53.0                                 & 37.3                                 & 50.4                                 & 71.3                                 & 72.3                                 & 47.6                                 & 15.9                                 & 50.8                                 & 63.6                                 & 60.8                                 & 52.3                                 \\
Restormer~\cite{zamir2022restormer}       & 53.8                                 & 43.8                                 & 51.4                                 & 68.7                                 & 66.8                                 & 52.6                                 & 21.6                                 & {\color[HTML]{00B0F0} \textbf{54.8}} & 59.8                                 & 63.3                                 & 53.7                                 \\
SCI~\cite{ma2022toward}             & 54.5                                 & 46.3                                 & 57.2                                 & 78.4                                 & 73.3                                 & 49.1                                 & 22.8                                 & 49.0                                 & 62.1                                 & 66.9                                 & 56.0                                 \\
SNR-Net~\cite{xu2022snr}         & {\color[HTML]{00B0F0} \textbf{57.7}} & {\color[HTML]{00B0F0} \textbf{48.6}} & {\color[HTML]{00B0F0} \textbf{59.5}} & {\color[HTML]{00B0F0} \textbf{81.3}} & {\color[HTML]{00B0F0} \textbf{74.8}} & 50.2                                 & {\color[HTML]{00B0F0} \textbf{24.4}} & 50.7                                 & {\color[HTML]{00B0F0} \textbf{64.3}} & 68.7                                 & {\color[HTML]{00B0F0} \textbf{58.0}} \\
Retformer~\cite{Cai2023}   & 50.9                                 & 47.7                                 & 58.6                                 & 77.2                                 & 68.1                                 & {\color[HTML]{00B0F0} \textbf{53.2}} & 17.4                                 & 52.0                                 & 61.3                                 & {\color[HTML]{00B0F0} \textbf{71.5}} & 55.8                                 \\
Ours & {\color[HTML]{FF0000} \textbf{59.8}} & {\color[HTML]{FF0000} \textbf{51.5}} & {\color[HTML]{FF0000} \textbf{62.1}} & {\color[HTML]{FF0000} \textbf{85.5}} & {\color[HTML]{FF0000} \textbf{76.6}} & {\color[HTML]{FF0000} \textbf{57.7}} & {\color[HTML]{FF0000} \textbf{28.9}} & {\color[HTML]{FF0000} \textbf{56.3}} & {\color[HTML]{FF0000} \textbf{66.2}} & {\color[HTML]{FF0000} \textbf{73.4}} & {\color[HTML]{FF0000} \textbf{61.8}} \\ \bottomrule
\end{tabular}}
\caption{Low-light semantic segmentation, where images are darkened by~\cite{zhang2021learning}. }
		\label{table:SemanticSegmentation} \vspace{-6mm}
\end{minipage}
\begin{minipage}[c]{0.432\textwidth}
\centering
\setlength{\abovecaptionskip}{0cm}
\resizebox{\columnwidth}{!}{
\setlength{\tabcolsep}{0.6mm}
\begin{tabular}{l|cccc|cccc}
\toprule
& \multicolumn{4}{c|}{\textit{COD10K}} & \multicolumn{4}{c}{\textit{NC4K}}   \\ \cline{2-9}
\multirow{-2}{*}{Methods} & \cellcolor{gray!40}$M$~$\downarrow$ & \cellcolor{gray!40}$F_\beta$~$\uparrow$ & \cellcolor{gray!40}$E_\phi$~$\uparrow$ & \cellcolor{gray!40}$S_\alpha$~$\uparrow$ & \cellcolor{gray!40}$M$~$\downarrow$ & \cellcolor{gray!40}$F_\beta$~$\uparrow$ & \cellcolor{gray!40}$E_\phi$~$\uparrow$ & \cellcolor{gray!40}$S_\alpha$~$\uparrow$ \\ \midrule
Baseline & 0.050 &0.625 &0.812 &0.756 & 0.071 & 0.733&0.816&0.763 \\
RetinexNet~\cite{wei2018deep}& 0.041                                 & 0.667                                 & 0.845                                 & 0.789                                 & 0.055                                 & 0.750                                 & 0.842                                 & 0.819                                 \\
KinD~\cite{zhang2019kindling}& 0.039                                 & 0.673                                 & 0.849                                 & 0.792                                 & 0.052                                 & 0.762                                 & 0.875                                 & 0.822                                 \\
MIRNet~\cite{zamir2020learning}                    & 0.037                                 & 0.697                                 & 0.857                                 & 0.799                                 & {\color[HTML]{00B0F0} \textbf{0.049}} & {\color[HTML]{00B0F0} \textbf{0.802}} & 0.888                                 & 0.833                                 \\
RUAS~\cite{liu2021retinex}& {\color[HTML]{00B0F0} \textbf{0.036}} & 0.705                                 & 0.861                                 & 0.803                                 & 0.051                                 & 0.795                                 & 0.883                                 & 0.827                                 \\
Restormer~\cite{zamir2022restormer}                 & {\color[HTML]{00B0F0} \textbf{0.036}} & 0.700                                 & 0.859                                 & 0.800                                 & 0.050                                 & 0.792                                 & 0.880                                 & 0.830                                 \\
SCI~\cite{ma2022toward}   & 0.037                                 & {\color[HTML]{00B0F0} \textbf{0.710}} & 0.863                                 & 0.805                                 & 0.051                                 & 0.782                                 & 0.880                                 & 0.836                                 \\
SNR-Net~\cite{xu2022snr} & {\color[HTML]{00B0F0} \textbf{0.036}} & 0.703                                 & {\color[HTML]{00B0F0} \textbf{0.865}} & 0.803                                 & {\color[HTML]{00B0F0} \textbf{0.049}} & 0.801                                 & {\color[HTML]{00B0F0} \textbf{0.892}} & {\color[HTML]{00B0F0} \textbf{0.838}} \\
Retformer~\cite{Cai2023}             & 0.037                                 & 0.682                                 & 0.861                                 & {\color[HTML]{00B0F0} \textbf{0.806}} & 0.052                                 & 0.766                                 & 0.881                                 & 0.832                                 \\
Ours   & {\color[HTML]{FF0000} \textbf{0.034}} & {\color[HTML]{FF0000} \textbf{0.725}} & {\color[HTML]{FF0000} \textbf{0.880}} & {\color[HTML]{FF0000} \textbf{0.813}} & {\color[HTML]{FF0000} \textbf{0.047}} & {\color[HTML]{FF0000} \textbf{0.804}} & {\color[HTML]{FF0000} \textbf{0.897}} & {\color[HTML]{FF0000} \textbf{0.841}} \\ \bottomrule
\end{tabular}}
\caption{Low-light concealed object segmentation.
}
		\label{table:CODSegmentation} \vspace{-6mm}
\end{minipage}
\end{table*}

\begin{figure}[t]
\begin{minipage}[c]{\textwidth}
	\centering
	\setlength{\abovecaptionskip}{-0.2cm}
	\begin{center}
		\includegraphics[width=\linewidth]{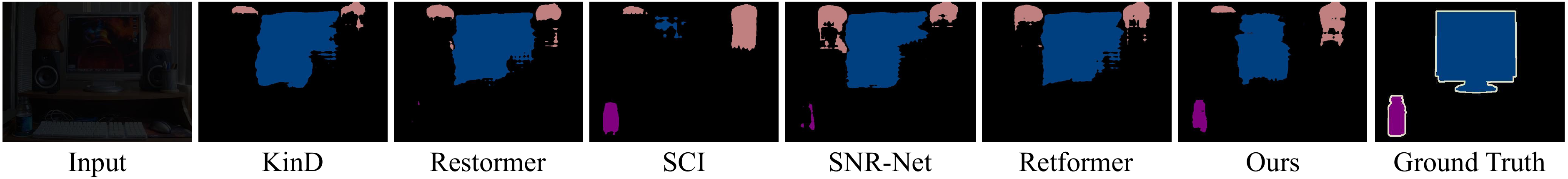}
	\end{center}
 \vspace{-1.3mm}
	\caption{Results on the low-light semantic segmentation task.}
	\label{fig:SemanticSegmentation}
\end{minipage} \\
\begin{minipage}[c]{\textwidth}
	\centering
	\setlength{\abovecaptionskip}{-0.2cm}
	\begin{center}
		\includegraphics[width=\linewidth]{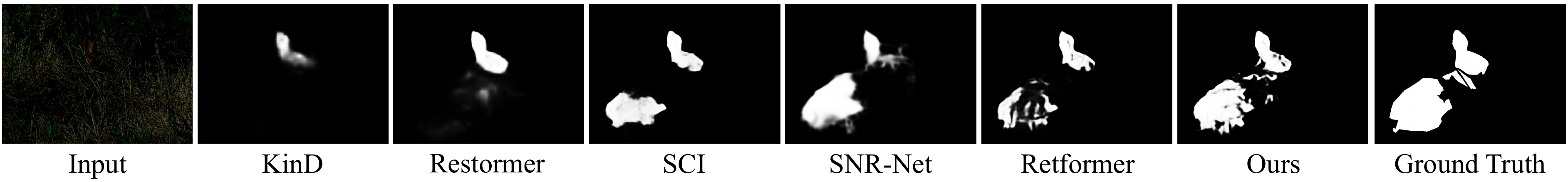}
	\end{center}
 \vspace{-1.3mm}
	\caption{Results on the low-light concealed object segmentation task.}
	\label{fig:COS}
	\vspace{-0.6cm}
\end{minipage}
\end{figure}

\subsection{User Study and Downstream Tasks}

\noindent\textbf{User Study.} We conduct a user study to assess the subjective visual perception of low-light image enhancement. In this study, 29 human subjects are invited to assign scores to the enhanced results based on four criteria: (1) The presence of underexposed or overexposed regions. (2) The existence of color distortion. (3) The occurrence of undesired noise or artifacts. (4) The inclusion of essential structural details. Participants rate the results on a scale from 1 (worst) to 5 (best). Each low-light image is presented alongside its enhanced results, with the names of the enhancement methods concealed.
The scores are reported in~\cref{table:UserStudy}, where our method receives the highest scores across all four datasets. This highlights our effectiveness in generating visually appealing results.

\noindent\textbf{Low-light Object Detection.} The enhanced images are expected to have better downstream performance than the original ones. We first verify this on low-light object detection. Following~\cite{Cai2023}, all compared methods are performed on \textit{ExDark}~\cite{loh2019getting} with YOLOv3, which is retrained from scratch with their corresponding enhanced results. As shown in~\cref{table:Detection}, our Reti-Diff exhibits a substantial advantage over existing methods
and the performance of our method surpasses that of the second-best method, Retformer~\cite{Cai2023}, by $4.72\%$,
which verifies
our efficacy in facilitating high-level vision understanding.

\noindent\textbf{Low-light Image Segmentation.} We extend our experiment to segmentation tasks, \ie, semantic segmentation and concealed object segmentation. Following the practice in detection, we also retrain the segmentor for each method.
This means that each method's enhanced results are segmented by the corresponding segmentor with specific weights. We argue this could better exploit the potential of image enhancement methods as a degraded data restoration module.

For semantic segmentation, following~\cite{ju2022sllen}, 
we apply image darkening to samples from the \textit{VOC}~\cite{everingham2010pascal} dataset according to~\cite{zhang2021learning}. We then employ  Mask2Former~\cite{cheng2022masked}
to perform segmentation on the enhanced results of these darkened images. 
We select Intersection over Union (IoU) for evaluation, and the results are presented in~\cref{table:SemanticSegmentation}. As shown in~\cref{table:SemanticSegmentation}, our method achieves the highest performance across all classes, surpassing the second-best method by 
$7.53\%$.

We further venture into concealed object segmentation (COS) on two widely-used datasets, \textit{COD10K}~\cite{fan2021concealed} and \textit{NC4K}~\cite{lv2021simultaneously}, which represents a challenging segmentation task aimed at delineating objects with inherent background similarity. We also apply image darkening~\cite{zhang2021learning} and enlist the cutting-edge COS segmentor, FEDER~\cite{he2023camouflaged}, to perform segmentation on the enhanced results. We evaluate the results using four metrics: mean absolute error $(M)$, adaptive F-measure $(F_\beta)$, mean E-measure $(E_\phi)$, and structure measure $(S_\alpha)$, which are presented in~\cref{table:CODSegmentation}. As depicted in~\cref{table:CODSegmentation}, our method exhibits superior performance compared to the second-best method, SNR-Net, with a margin of $2.16\%$ on average. Note that it is a notable improvement in COS. Collectively, the exceptional results achieved in these two segmentation tasks substantiate our proficiency in recovering image-level illumination degraded information.

\begin{figure*}[t]
	\centering
	\setlength{\abovecaptionskip}{-0.2cm}
	\begin{center}
		\includegraphics[width=\linewidth]{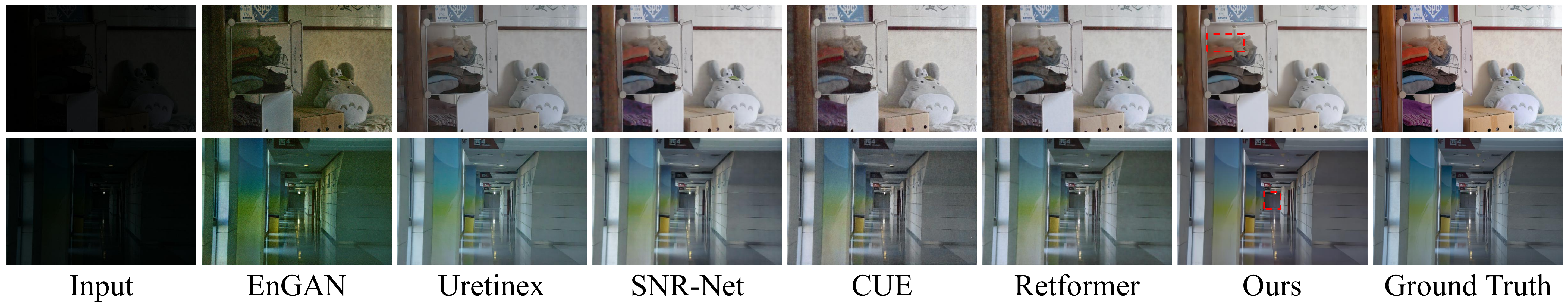}
	\end{center}
 \vspace{-1.3mm}
	\caption{Failure cases. Our results show blurred texture details in the dashed boxes.}
	\label{fig:limitations}
	\vspace{-0.6cm}
\end{figure*}

\section{Discussions}\vspace{-2mm}
Our Reti-Diff is the first LDM-based solution specifically designed for the IDIR task, setting it apart from existing LDM-based methods applied in other tasks. To illustrate the distinctions, we compare it with a general enhancement method, DiffIR~\cite{xia2023diffir}: \textbf{(1) Motivation.} Reti-Diff targets enhancing details and correcting degraded illumination. Thus, we enable RLDM to learn Retinex knowledge and generate Retinex priors from the low-quality input. We contend that relying solely on priors extracted from the RGB domain struggles to fully represent valuable texture details and correct illumination cues, leading to suboptimal restoration performance. To verify this, we substitute our RLDM for the LDM structure used in DiffIR. In \textit{LOL-v2-syn}, we observe that the PSNR rises from 24.76 to 26.14 and the SSIM increases from 0.921 to 0.933.
\textbf{(2) Implementation.} Apart from proposing RLDM to extract Retinex priors, we further modify the structure of RGformer to implicitly model the Retinex theory at the feature level and introduce an auxiliary decoder to reconstruct the decomposed Retinex components to the RGB domain.
\textbf{(3) Performance.} As shown in~\cref{table:low-light}, our Reti-Diff significantly outperforms DiffIR~\cite{xia2023diffir} by $20.6\%$ on average.
\vspace{-2mm}
\section{Limitations and Future Work}\vspace{-2mm}
As shown in~\cref{fig:limitations}, our Reti-Diff encounters challenges in simultaneously recovering illumination and restoring texture details when the low-quality inputs exhibit severe illumination degradation. This issue persists across existing methods and remains unresolved. We attribute this to the loss of texture information during illumination recovery. To address this limitation in future research, we propose excavating texture priors from other domains, \eg, the frequency domain. These priors can complement the reflectance priors extracted from the RGB domain, enhancing the preservation of critical texture features. 
Additionally, we consider the use of multimodal data~\cite{fang2023joint} to aid in improving image reconstruction performance, such as using infrared images~\cite{he2023degradation,xu2023dm,ju2022ivf} to aid in low-light visible image enhancement. Besides, We will explore whether our approach is downstream task-friendly with more segmentation algorithms~\cite{he2023weakly,he2019image,xiao2023concealed}.
We also aim to extend our approach to tackle IDIR problems afflicted by other types of degradation, such as haze and motion blur, using some domain adaptation strategies~\cite{tang2023consistency,tang2023source}. These endeavors will further advance the capabilities and applicability of Reti-Diff in real-world scenarios.

\vspace{-2mm}
\section{Conclusions}\vspace{-2mm}
To balance generation capability and computational efficiency,
our approach adopts DM within a compact latent space to generate guidance priors. Specifically, we introduce RLDM to extract Retinex priors, which are subsequently supplied to RGformer for feature decomposition. This process ensures precise detailed reconstruction and effective illumination correction. RGformer then refines and aggregates the decomposed features, enhancing the robustness in handling complex degradation scenarios. Our approach is extensively validated through experiments, establishing the clear superiority of the proposed Reti-Diff.

\clearpage  

%
%
\bibliographystyle{splncs04}
\bibliography{egbib}
\end{document}